%% file: main_arxiv.tex
\pdfoutput=1

\documentclass[11pt]{article}

\usepackage[preprint]{acl}

\usepackage{times}
\usepackage{latexsym}
\usepackage{algorithm}    
\usepackage{algpseudocode}
\usepackage{bm}
\usepackage{soul} 

\usepackage{enumitem}

\usepackage[table]{xcolor}
\usepackage{multirow}

\usepackage{rotating}

\usepackage{pifont}
\newcommand{\cmark}{\ding{51}}
\newcommand{\xmark}{\ding{55}}

\usepackage[T1]{fontenc}

\usepackage[utf8]{inputenc}

\usepackage{microtype}

\usepackage{inconsolata}

\usepackage{graphicx}

\usepackage{xcolor}
\usepackage{booktabs}
\usepackage[most]{tcolorbox} 
\usepackage{multirow}
\usepackage{xspace}

\usepackage[normalem]{ulem}

\def\myalgo{\textsf{Luminol-AIDetect}\xspace}
\def\ppl{\mathit{ppl}}
\def\pplshuf{\mathit{ppl}_{\mathit{shuf}}}
\newcommand{\hgt}{\textsc{hgt}\xspace}
\newcommand{\mgt}{\textsc{mgt}\xspace}

 \title{Luminol-AIDetect: Fast  Zero-shot  Machine-Generated Text Detection\\ based on Perplexity under Text Shuffling}

\author{Lucio La Cava, Andrea Tagarelli \\
  DIMES Dept., University of Calabria \\
  v. P. Bucci 44Z, 87036 Rende, CS, Italy\\ 
  \texttt{\{lucio.lacava,tagarelli\}@dimes.unical.it} \\ 
  }

\begin{document}

\maketitle

\begin{abstract} 
Machine-generated text (\mgt)  detection requires identifying structurally invariant signals across generation models, rather than relying on model-specific fingerprints. 
In this respect, we hypothesize that while large language models excel at local semantic consistency, their autoregressive nature results in a specific kind of structural fragility compared to human writing. 
We propose \myalgo, a novel,   zero-shot  statistical approach that exposes this fragility through coherence disruption. 
By applying a simple randomized text-shuffling procedure, we demonstrate that the resulting shift in perplexity serves as a principled, model-agnostic discriminant, as \mgt displays a characteristic dispersion in perplexity-under-shuffling that differs markedly from the more stable structural variability of human-written text. \myalgo leverages this distinction to inform its decision process, 
where a handful of  perplexity-based scalar features are extracted from an input text and its shuffled version, 
then  detection is performed via density estimation and ensemble-based prediction. 
Evaluated  across 8 content domains, 11 adversarial attack types, and 18 languages, \myalgo demonstrates state-of-the-art performance, with gains up to 17x lower FPR while being cheaper  than prior methods. 

\end{abstract}

\section{Introduction}
Large language models (LLMs) can nowadays 
produce fluent translations, coherent summaries,   accurate question answering, and more,  across different domains and languages. These capabilities had a striking impact on communication, making LLMs an equalizer of information access. Beyond individual productivity~\cite{science-productivity}, LLMs are concretely accelerating scientific progress~\cite{ren2025towards,zhang-etal-2024-comprehensive-survey}, education~\cite{vajjala2025opportunities}, and software development~\cite{jiang2026code}, reaching unprecedented scale. 
However, the human-like generative capabilities of LLMs also render them susceptible to misuse. Key concerns include hard-to-detect LLM-generated misinformation~\cite{chen2024can} and fake news~\cite{hu2025fakenews}, the growing adoption of LLM-generated content in critical domains such as academia~\cite{liang2024mapping, russo2025lottery}---raising significant concerns regarding transparency, authorship, and the integrity of scientific knowledge---and an increasing adoption in social media platforms~\cite{la2025machines}.

All these scenarios share a single point: LLM-generated text, hereinafter referred to as \textit{machine-generated text} (\mgt), and \textit{human-generated text} (\hgt) are nowadays often \textit{indistinguishable} to human evaluators, effectively serving as a modern-day demonstration of the Turing Test in practice.

To this aim, researchers have been developing approaches for distinguishing between \hgt and \mgt~\cite{jawahar-etal-2020-automatic,wu-etal-2025-survey, tang2024science}. However, as detector capabilities improve, LLMs do too. This dynamic co-evolution implies that each side continuously adapts in response to the other: detectors are increasingly precise, while LLMs keep improving alignment with human styles, rendering detectors potentially obsolete tomorrow. 
This observation has a key implication which motivates our work, also in line with current statistical approaches for \mgt detection:  \textit{robust detection should require to capture signals in a \mgt   that are structurally invariant to the text-generation model, rather than a learned model-specific fingerprint}.

\begin{figure}[t]
    \centering
    \includegraphics[width=\linewidth]{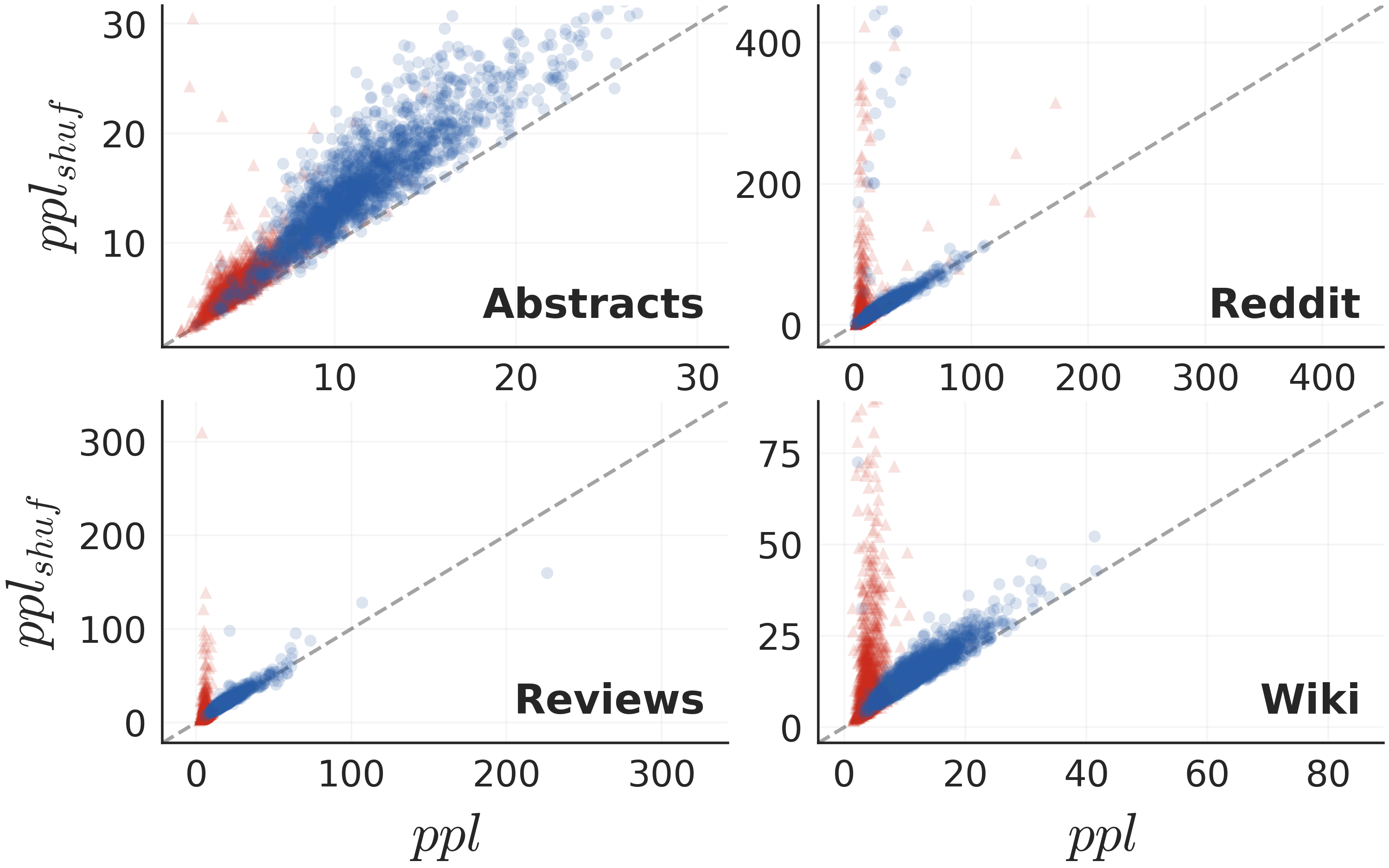}
    \caption{Text vs. shuffled-text perplexities on \hgt (blue) and \mgt (red) instances, for various domains.}
    \label{fig:motivating}
\end{figure}

\vspace{1mm}
\noindent\textbf{Our hypothesis.\ }
 Decoder  LLMs generate text autoregressively, so that each
 token is sampled from a probability distribution 
 produced by a softmax bottleneck. 
 LLMs excel at maintaining local semantic consistency, however, because the sampling process is a greedy or locally-stochastic heuristic, they lack a global planning objective,  
so there is no guarantee of globally optimal or uniquely ``most likely'' sentence-level continuations.   
By contrast, human writing, being driven by communicative intent, exhibits higher variability in transitions, 
which results in greater lexical diversity 
and long-range structural variance that autoregressive models often smooth 
away into a statistical average.
 
Building on this remark, our hypothesis is that the inherent difference of sequential-structure nature and strength between \hgt and \mgt  might be 
  exposed by a simple procedure that breaks down the text  coherence, 
then this  signal can be encoded into \textit{perplexity-based scalar-features}. 
We expect that \textit{the higher impact in terms of coherence disruption exerted by the  text-breaking on \mgt than on \hgt can be captured by such perplexity-features and then exploited  as a \mgt detection clue.} 

To this purpose, we started by  defining the  text-breaking procedure. Our intuition is suggestive: a simple randomized procedure like \textit{text shuffling} suffices. 
To support this, let's consider Fig.~\ref{fig:motivating} which  plots the text perplexities (x-axis)   versus shuffled-text perplexities (y-axis) for a set of \mgt and \hgt instances:   
although the behavior may change across domains,  the  perplexities on \mgt display a higher concentration before shuffling and a larger dispersion after shuffling compared to \hgt.

\vspace{2mm}
\noindent\textbf{Contributions.\ }
We propose \myalgo, a novel statistical approach to \mgt detection   that:

\vspace{-3mm}
\begin{itemize}[itemsep=-2pt, leftmargin=*]
    \item Leverages the \textit{perplexity-under-shuffling} signal as a statistically principled, text-generator-agnostic discriminant between \mgt and \hgt, based on a handful  of numerical features derived from the perplexities on an input text and its shuffled version, by  a small  decoder LLM;  
    \item Makes no assumption on which LLM generated the text, thus being applicable in all real-world deployment scenarios, including closed-source, API-only, and unknown-provenance settings;
    \item Is 
    zero-shot as it only requires to  estimating parametric density functions, based on   probability distributions \textit{once-only}  fitted to   perplexity-feature data samples.  
    \item Thus, at inference time, \myalgo just requires to shuffle the input text and   to lookup for their  perplexity-features into probability density values to predict if the input text is \mgt.  
    This is hence  substantially cheaper than other black-box perturbation-based methods such as the state-of-the-art Binoculars~\cite{hans2024binoculars} and  Fast-DetectGPT~\cite{bao2024fastdetectgpt}.
\end{itemize}

\myalgo's effectiveness is demonstrated through extensive evaluation on the widely used RAID and MULTITuDE benchmarks, which  jointly capture   key dimensions of real-world \mgt  detection, spanning diverse  domains, \mgt models, adversarial attacks, and  languages.

\section{Related work}
Nowadays, most efforts to detect whether a text is \mgt or \hgt are developing across two main categories: \textit{model-based} and \textit{metric-based} approaches.

The former include deep learning frameworks for detecting \mgt~\cite{verma-etal-2024-ghostbuster,Bhattacharjee2024gpt,Uchendu24}, along with contrastive learning approaches~\cite{la-cava-tagarelli-2025-openturingbench,LaCava2024whosai, bhattacharjee-etal-2023-conda}, adversarial training~\cite{bhattacharjee2024eagle}, topological aspects~\cite{Uchendu24}, and discourse motifs~\cite{kim-etal-2024-threads}.

The latter category, which is where our proposed approach focuses on, includes approaches that leverage statistical, token-level  probabilities and statistical methods to determine whether a text has been machine-generated, e.g., by using rank-related scores~\cite{su-etal-2023-detectllm}, entropy~\cite{gehrmann-etal-2019-gltr}, perplexity~\cite{wang-etal-2023-seqxgpt}, information density~\cite{venkatraman-etal-2024-gpt}.  
Among these, the most widely adopted detectors exploit the impact of text manipulation (e.g., via perturbation) on model-assigned probabilities as a detection signal. Notable examples are DetectGPT~\cite{pmlr-v202-mitchell23a} and Fast-DetectGPT~\cite{bao2024fastdetectgpt}, which perturb the candidate text with a masked language model and measures whether the generating LLM's log-probability systematically decreases, a pattern that holds for \mgt  (which sits near local probability maxima) but not for human-generated one. 
Binoculars~\cite{hans2024binoculars} computes a cross-model likelihood ratio between a scorer and a reference model,  measuring how anomalously one model rates the text relative to another.

However, current metric-based approaches require generating multiple variations, depend on multiple models during inference, and the transformations they apply (stochastic token-level resampling or cross-model scoring) produce a signal that varies with the choice of surrogate. Similarly, deep learning frameworks are inherently generator-specific: trained on samples from a fixed set of LLMs, they learn stylistic and distributional fingerprints that might not transfer to different generators.

\section{The \myalgo method}
\begin{figure}
    \centering
    \includegraphics[width=\linewidth]{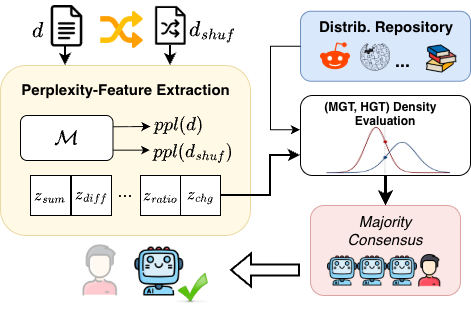}
    \caption{\myalgo  at inference.}
    \label{fig:workflow}
\end{figure}

 Figure \ref{fig:workflow} provides an  overview of  our \myalgo, whose workflow consists of four main phases: (1) text shuffling, 
 (2) computation of perplexities and extraction of perplexity-features, 
 (3) domain-specific data distribution   fitting or evaluation, 
 (4), ensemble-based \mgt prediction.  

Phases 1–3 are performed only once for a given collection of texts within a specific domain. At inference time, each query text   requires fast  processing  through Phases 1, 2, and 4.  The following sections describe  each phase in   detail.

\subsection{Text shuffling}
\label{sec:shuffling}

Let $d = \langle P_1,\ldots,P_m \rangle$ be a text with     
$m \geq 1$ paragraphs, where each $P_i = \langle s_{i,1},\ldots,s_{i,n_i}\rangle$
contains $n_i \geq 1$ sentences. Let  $\pi$ denote    a uniformly random permutation of word positions, and   $\sigma$ a uniformly random permutation of sentence positions.
We define a simple procedure of text shuffling for $d$, whose output  $d_{\mathit{shuf}}$ 
is produced as follows:  

\noindent 
$\bullet$
 If $d$ is a single-sentence paragraph $P = \langle s\rangle$, with $s$  
containing $k$ words, word-level shuffling is carried out: for any $s = w_1 \cdots w_k \cdot p$, where $w_1,\ldots,w_k$ are the $k$ words and $p$ is the terminal punctuation mark,  $d_{\mathit{shuf}} = \pi(w_1,\ldots,w_k) \cdot p$; 

\noindent 
$\bullet$ If $d$ is a  multiple-sentence paragraph,  $P_1$, 
then  $d_{\mathit{shuf}} =  \sigma(P_1) = \langle s_{1,\sigma(1)},\ldots,s_{1,\sigma(n_1)}\rangle$;  

 \noindent 
$\bullet$ If $d$ is multi-paragraph ($m>1$), the above single-paragraph procedure is applied to each $P_i$ independently, and the results are concatenated with  their original paragraph order, to produce  $d_{\mathit{shuf}}$.

Note that the above presented procedure should be intended as a simple proof-of-concept   for a ``least-effort'' text shuffling that operates under minimal structural disruption. Our text shuffling method is in fact  single-pass, paragraph-order preserving (thus avoiding breaking the  higher-level discourse organization), and sentence-level shuffling in all cases, except for   single-sentence texts, in which case words are shuffled. 
Clearly, alternative, more refined shuffling procedures could be used, which are left for future  investigation; however, the presented  approach suffices for our purposes.

\subsection{Perplexity feature extraction}
\label{sec:extraction}

\paragraph{Perplexity.}
For both the original text ($d$) and its shuffled version ($d_{\mathit{shuf}}$), we compute the perplexity by using a single, small, open-weight  proxy model $\mathcal{M}$. 
For any tokenized sequence $(w_1,...,w_T)$,  the perplexity of $\mathcal{M}$ on $d$ is defined as 
$perplexity(\mathcal{M},d)=\exp \left(\frac{1}{|N|} \sum_{i\in N}\ell_i \right)
$
where $\ell_i = -\log \Pr_\mathcal{M}
(w_i|w_{<i})$, $N$ is the set of target positions $i>W$ considering sliding window of context size $W$ (model max length), stride $S$ (typically,   $S=W/2$), with each token predicted exactly once across the document. 
Note  that $\mathcal{M}$ is entirely \textit{independent} from the model that generated $d$. 
Hereinafter, we will use $\ppl$ to denote $perplexity(\mathcal{M},d)$ and $\pplshuf$ to denote $perplexity(\mathcal{M},d_{\mathit{shuf}})$.

\paragraph{Perplexity features.}
For any document $d$, and its associated shuffled version $d_{\mathit{shuf}}$, the corresponding pair $(\ppl, \pplshuf)$ is expected to encode the  paired signal supporting  our hypothesis of  statistical difference in perplexity between a text and its shuffled version.  
To unveil this, we consider   different ways of turning a pair $(\ppl, \pplshuf)$ into scalar \textit{features} that capture contrast and scale in a simple, interpretable manner. 
More specifically, we want to account for (i) total magnitude, (ii) absolute magnitude difference, (iii)
relative change, (iv) scaled/symmetric  relative change, and (v) relative change as a percentage.  
Note that we do not require   these properties   to be captured by a single statistic; 
rather, we seek to analyze each separately to maintain their distinct contributions, through the following operations: \textit{sum}, \textit{difference}, \textit{ratio}, \textit{log-ratio}, and \textit{percent change}. 
For short, we will use symbol  $\phi$ to denote  a \textit{perplexity-feature type} in the set $\Phi$ of the above operators, 
and $z_{\phi}$ to denote a \textit{perplexity-feature} of  type $\phi$. For instance,  $z_{\mathit{diff}}=\pplshuf - \ppl$, or $z_{\mathit{ratio}}  =\pplshuf / \ppl$.

\paragraph{(\mgt, \hgt)-paired feature significance analysis.}
The perplexity-feature types above presented are hence conceived as measures of the combination or deviation   of the perplexity on a shuffled text with the perplexity on the original text. Beyond this,  we would like to   exploit them for a higher purpose, i.e., to discriminate between   \mgt and \hgt. This raises a preliminary yet essential question:  
 given  a domain $\mathcal{D}$, and    any perplexity-feature type $\phi$, whether the values $z_{\phi}$ computed  for  \hgt instances are statistically different from the  $z_{\phi}$ computed  for \mgt instances that are ``paired'' with  the former.  

To answer the above question,  we considered \mgt-\hgt parallel data  (e.g.,  benchmark corpora containing \hgt instances and associated \mgt instances),  for   various domains $\mathcal{D}$. Given a choice for $\mathcal{D}$, and for each perplexity-feature type $\phi$, we assessed the statistically significant difference in the mean values (Welch’s t-test), or in the location/median shift (Mann–Whitney U test) 
between the   \hgt's $z_{\phi}$ observations and the   \mgt's $z_{\phi}$ observations. 
Our experimental results, conducted across a range of domains (cf. \uline{Appendix} \ref{appx:stats}),  confirmed statistical evidence that the \mgt and \hgt $z_{\phi}$ observations, by varying feature type $\phi$, indeed differ, according to the tests' outcomes.

\subsection{Perplexity feature distribution fitting}
\label{sec:fitting}

One obtained empirical evidence of the \mgt/\hgt discriminating role that the perplexity  features can play, 
we  pursue our goal of   defining a  method for checking if a text (with associated   shuffled version)  represented by some perplexity feature $z_{\phi}$, is most likely from \mgt or \hgt probability  distributions.

\paragraph{Bootstrapping. }
First, we investigate 
which continuous distribution families are likely to be used,  to properly fit  the   \mgt's and \hgt's $z_{\phi}$ observations.

To this purpose, we aim to  fit parametric continuous  distribution families (cf. \uline{Appendix}~\ref{appx:details} for the full list of families used) to  data selected from any $\mathcal{D}$ which correspond to  each of the     $z_\phi$ observations, for both \hgt and \mgt independently, and assess goodness of fit using a \textit{bootstrap} Kolmogorov–Smirnov (KS) test, where distribution parameters are estimated from the data, under a one-sample setting (i.e., data vs. a single theoretical distribution).
Higher bootstrap p-values indicate better fit; p-values exceeding the significance threshold (e.g., 0.05) indicate that there is insufficient evidence to reject the null hypothesis that the data follow the specified distribution. 

It should be emphasized that the bootstrap KS test is   performed \textit{one-time only}. Also, though it  is specific for  any $z_\phi$  data sample of a given domain $\mathcal{D}$, the outcome of the test can be generalized to new data  from the same domain.

\paragraph{Model selection and fitting.}
We are now given a set   $\mathbf{F}_{\phi}$  of continuous distribution families for which the bootstrap KS test applied to each set of  $z_{\phi}$ observations failed to  reject the null hypothesis  \textit{for both \mgt and \hgt } observations (from a given domain $\mathcal{D}$). 
This set can contain one or more distribution families, or even none in principle, although our selection of distribution families to test was large enough to avoid this extreme case. 
Thus, for any (non-empty) set $\mathbf{F}_{\phi}$, we retain only the best-fitting  distribution family, i.e., the one  that corresponded to the higher bootstrap p-value. 
Let us   denote this best-fitting distribution family with   $\mathcal{F}_{\phi}$, and the set of perplexity-feature types having (at least) one fitting distribution  with  $\hat{\bm{\Phi}}$.

The best-fitting distribution families, along with their MLE parameters, are eventually stored in a fitted distribution repository, we denote with $\Omega$,  associated with each domain $\mathcal{D}$, i.e.,   
\begin{equation}
\hspace{-2.3mm}
\Omega\!=\!\left\{
  \bigl\langle {\mathcal{F}}^{(c)}_{\phi},\; \theta^{(c)}_{\phi} \bigr\rangle
  \!:\! \phi \in \hat{\bm{\Phi}},
  c \in \{\mgt,\hgt\}
\right\}\!\!.
\label{eq:repo}
\end{equation}

Notably, these domain-specific repositories represent the \textit{only} fitted artifacts across our entire framework, each yielding a compact $(2|\hat{\bm{\Phi}}|)$ parameter vector.   They also offer the key advantage of statistical interpretability as well as of being agnostic to the  choice of  text-generation model.

\subsection{Inference}
\label{sed:inference}

Here we describe how \myalgo   works at  inference time, as reported in Algorithm~\ref{alg:myalgo}.    
Given a query  text $d$, the algorithm outputs the predicted  probability  for the class \mgt, through the following main steps which are detailed next.  

\paragraph{Density evaluation.}
After shuffling the query text (Line 1) and computing the two texts' perplexities (Lines 2-3), \myalgo first loads the  $\mathcal{D}$-specific $\Omega$ repository, 
along with the set $\hat{\bm{\Phi}}$ of perplexity-feature types associated with 
$\mathcal{D}$  (Lines 4-5).
 Then, it iterates over  the feature-types (Lines 7-15) accessing the fitting distribution family ${\mathcal{F}}^{(c)}_{\phi}$ and its parameters $\theta^{(c)}_{\phi}$, for each feature $\phi \in \hat{\bm{\Phi}}$ and each class $c \in \{\mgt, \hgt\}$, in order  to evaluate the corresponding probability density functions at each observed
feature value $z_\phi$  (Lines 10-11): 
\begin{align}
  pdf^{(\mgt)}_{z_\phi} &= {\mathcal{F}}^{(\mgt)}_{\phi}.\mathit{pdf}
                  \!\left(z_\phi;\;\theta^{(\mgt)}_{\phi}\right), \\
  pdf^{(\hgt)}_{z_\phi} &= {\mathcal{F}}^{(\hgt)}_{\phi}.\mathit{pdf}
                  \!\left(z_\phi;\;\theta^{(\hgt)}_{\phi}\right).
\end{align}

\begin{algorithm}[t!]
\caption{\myalgo}
\label{alg:myalgo}
\small
\begin{algorithmic}[1]
\makeatletter
\renewcommand{\Comment}[1]{{\color{teal}$\triangleright$ \ \texttt{#1}}}
\makeatother
\Require a text $d$ of domain $\mathcal{D}$
\Require   a decoder LLM $\mathcal{M}$ (default:  \textit{GPTNeo});   
decision logic $\ell$ (default: \textit{`majority consensus'})

\Ensure probability that $d$ is \mgt

\State $d_{shuf} \gets \textsl{shuffling}(d)$ 
\Statex \Comment{Compute the perplexity for the original and shuffled texts}
\State $ppl \gets \textsl{perplexity}(\mathcal{M}, d)$
\State $ppl_{shuf} \gets \textsl{perplexity}(\mathcal{M}, d_{shuf})$
\Statex \Comment{Load the fitted distribution repository associated with $\mathcal{D}$}
\State $\Omega \gets getDistrRepo(\mathcal{D})$
\Statex \Comment{Load the perplexity-feature types associated with $\mathcal{D}$}
\State $\hat{\bm{\Phi}} \gets getFeatureTypes(\mathcal{D})$
\Statex \Comment{Loop~over~the~feature-types}
\State $\mathcal{E} = \{\}$ \quad\Comment{Ensemble of (M,H)-densities}
\For{$\phi \in \hat{\bm{\Phi}}$}   
\State  $z_{\phi} \gets getFeature(ppl, ppl_{shuf}, \phi)$
\State   $\langle \mathcal{F}, \theta\rangle^{(M)}, \langle\mathcal{F}, \theta\rangle^{(H)}\!~\gets~\!getParams(\Omega, \phi)$
\Statex \hspace{5mm} \Comment{Eval. the (M,H)-densities at $z_{\phi}$} \hfill
\State  $pdf_{z_{\phi}}^{(M)} \gets \mathcal{F}^{(M)}.pdf(z_{\phi}, \theta^{(M)})  $ 
\State $pdf_{z_{\phi}}^{(H)} \gets \mathcal{F}^{(H)}.pdf(z_{\phi}, \theta^{(H)})  $ 
 \If{{\bf not} \textsl{implausibility}($pdf_{z_{\phi}}^{(M)}, pdf_{z_{\phi}}^{(H)}$)}
\State $\mathcal{E}.add((pdf_{z_{\phi}}^{(M)}, pdf_{z_{\phi}}^{(H)}))$ 
\EndIf 
\EndFor
\Statex \Comment{Classify $d$}
\If{$\mathcal{E}$ is empty}
\Return $reject$ 
\Else
\State $ensemble\_probs \gets \left\{\frac{pdf_{z_{\phi}}^{(M)}}{pdf_{z_{\phi}}^{(M)} + pdf_{z_{\phi}}^{(H)}} \right\}_{\mathcal{E}}$
\State \Return $\textsl{decision}(ensemble\_probs, \ell)$
\EndIf
\end{algorithmic}
\end{algorithm}

\paragraph{Implausibility check.}
If both the density values $pdf^{(\mgt)}_{z_\phi}$ and $pdf^{(\hgt)}_{z_\phi}$ of the fitted \mgt- and \hgt-distributions, for any given perplexity-feature $z_{\phi}$, are extremely low, then they cannot be regarded as plausible, and hence we should take the decision of \textit{reject as neither distribution}. 
To this purpose, we define a simple  \textit{implausibility condition} check (cf. Line 12 of Algorithm \ref{alg:myalgo}), as: 
\begin{equation}
\label{eq:plausibility}
\max(pdf_{z_{\phi}}^{(c)}) < \tau,
\end{equation}
with $c \in \{\mgt, \hgt\}$. 
A principled choice for $\tau$ is based on quantiles over the observed data, i.e., 
$
\tau_c = \text{Quantile}_\alpha(pdf_{z_{\phi}}^{(c)}).
$    
A conservative choice is setting $\alpha=0.01$ (1st percentile), i.e., 1\% of false rejection. 
By using as $\tau = \min(\tau_c)$, we ensure that both \mgt and \hgt distributions retain at least $(1-\alpha)$\% of the  mass. 

It should be noted that if a value (i.e., perplexity-feature $z_{\phi}$) is found to be out-of-distribution (OOD) based on Eq. (\ref{eq:plausibility}), then we cannot decide about assignment to \mgt or \hgt, otherwise probabilistic guarantees are broken. 
On the other hand, OOD rejection is informative per se as it can have various interpretations. In particular, for a Burr type distribution, extreme right tail would correspond to unusually repetitive or verbose text, extreme left tail to very  short or compressed text, and likelihood valley to a domain mixture or domain shift.

\paragraph{Decision logic.}  
Let 
$\mathcal{E} = \{\hat{p}^{(\mgt)}_{z_\phi}\}$ 
denote the  \textit{ensemble of  \mgt prediction probabilities} $\hat{p}^{(\mgt)}_{z_\phi} = pdf^{(\mgt)}_{z_\phi} / (pdf^{(\mgt)}_{z_\phi} + pdf^{(\hgt)}_{z_\phi})$; if the implausibility check is enabled, we     retain  only  those feature-specific  densities that fail the implausibility test.  
Each feature type  $\phi$ contributes to $\mathcal{E}$ with a   vote:
\begin{equation}
v_\phi =
\begin{cases}
\mgt & \text{if } \hat{p}^{(\mgt)}_{z_\phi} \geq \hat{p}^{(\hgt)}_{z_\phi} - \Upsilon \\
\hgt & \text{otherwise.}
\end{cases}
\end{equation}

The final decision corresponds to the majority class $c^* = \arg\max_{c}\,|\{\phi : v_\phi = c\}|$: this prioritizes robustness, as a minority predictor (feature)  is unlikely to reverse the decision. 
Note that $\Upsilon$, which is set to 0 by default, can be used as an \textit{uncertainty threshold} to introduce a small tolerance coefficient into the decision in favor of \mgt. 
Details on the settings of $\Upsilon$ are in \uline{Appendix}~\ref{appx:details}.

\section{Experimental Setup}
\label{sec:setup}

\paragraph{\bf Evaluation Scenarios and Benchmarks.}
We consider the evaluation of   \mgt detection methods  across diverse real-world scenarios, focusing on the following tasks: 
\textbf{multi-domain} evaluation \textbf{(E1)}, 
robustness to \textbf{adversarial attacks (E2)}, and 
\textbf{multilingual/cross-lingual} evaluation \textbf{(E3)}. 
 
Task E1 is aimed  to assess a detector's performance on \mgt and \hgt of different domains (e.g., news, reviews, abstracts, social posts) where \mgt is produced by diverse LLMs.
Task E2  evaluates a detector on \mgt  altered through attacks of different types (e.g., paraphrasing, paragraph insertion, character-/word-level substitutions) attacks). 
Task E3   evaluates a detector   across multiple languages, focusing on the challenging   setting where the test languages are mostly different from the source language (English) on which the detector is based.

To support the above evaluation tasks, we resort to two widely-adopted benchmarks that together cover the key aspects relevant to real-world \mgt detection, namely \textit{content domains}, \textit{generating models}, \textit{adversarial attacks}, and \textit{source languages}. 
\textbf{RAID}~\cite{dugan-etal-2024-raid} is the most comprehensive   benchmark for \mgt detection, as it covers 8   domains---\textit{Abstracts}, \textit{Books}, \textit{News}, \textit{Poetry}, \textit{Recipes}, \textit{Reddit}, Reviews, and \textit{Wiki}  articles, with \hgt and \mgt instances---from 11 generators---varying in length starting from an average of 100 words (e.g., Abstracts or Reddit). It also includes 11 post-hoc adversarial attacks. 
RAID underpins the \myalgo\ instance used for the experimental evaluation of tasks E1 and E2.

For task E3, we resort to \textbf{MULTITuDE} (v3)~\cite{macko-etal-2023-multitude,macko_2025_15519413},  
a multilingual benchmark of \mgt and \hgt news articles  in 18 languages. 
Details can be found in \uline{Appendix}~\ref{appx:data}.

\vspace{-1mm}
\paragraph{\bf Competing Methods.}
We compared \myalgo to   state-of-the-art zero-shot approaches,   \textit{Fast-DetectGPT}~\cite{bao2024fastdetectgpt} and \textit{Binoculars}~\cite{hans2024binoculars}. 
We also  resorted to   representative  metric-based detectors, available in MGTBench~\cite{He2024mgtbench}: 
\textit{Log-Likelihood}~\cite{solaiman2019release}, 
\textit{Rank}, \textit{Entropy}, \textit{GLTR}~\cite{gehrmann-etal-2019-gltr}, 
\textit{Log-Rank}~\cite{pmlr-v202-mitchell23a},  
\textit{LRR}~\cite{su-etal-2023-detectllm};  
the latter   methods extract features that are fed to  a logistic  regressor.

\vspace{-1mm}
\paragraph{\bf Assessment Criteria.} 
Following~\cite{hans2024binoculars}, we evaluate the detectors based on \textit{false positives} (\hgt instances incorrectly classified as \mgt) and \textit{false negatives} (\mgt instances incorrectly classified as \hgt).
These criteria relate to the trade-off between sensitivity and specificity, and allow for  a 
practically meaningful assessment of a \mgt 
detector by reflecting critical deployment risks. 
For example, the false positive rate concerns \mgt evading detection (e.g., bypassing academic integrity checks or appearing as genuine reviews), whereas the false negative rate reflects legitimate human content being misclassified as \mgt.

\vspace{-1mm}
\paragraph{\bf \myalgo settings.}
To set $\mathcal{M}$, we opted for the commonly used GPT-Neo 2.7B (with $W=2048, S=1024$). Note,   however,  that any choice of $\mathcal{M}$ remains  totally  independent from the \mgt originating model.   \ 
For the bootstrapping, we evaluated 11 distribution  families; across all RAID domains, the best-fitting models were consistently the Burr and Gamma distributions (\uline{Appendix}~\ref{appx:settings}).

In the evaluation pipeline, we considered: (i) the option to disable the implausibility check; (ii) the application of outlier removal as a preprocessing step; and (iii) multiple settings of the  uncertainty threshold. See  \uline{Appendix}~\ref{appx:details} for details on (ii)-(iii).

\vspace{-1mm}
\paragraph{\bf Computational cost of \myalgo.}
Methods like Binoculars and Fast-DetectGPT require two forward passes plus a full vocabulary scan at each token. In contrast, \myalgo avoids the full-vocabulary step and operates only over observed tokens, reducing substantially the cost, which can additionally benefit from a 3x smaller proxy model than the other methods.  
We refer to  \uline{Appendix}~\ref{appx:costs} for details.

\vspace{-1mm}
\section{Results}
\vspace{-1mm}
  
\subsection{Perplexity-feature statistical analysis}
We first examined    the effect of shuffling one the   perplexities on \mgt and \hgt, and on  their associated perplexity-features. 
Overall, the higher impact of shuffling on \mgt than on \hgt is noted in the typically larger values of perplexity-features  in \mgt w.r.t. \hgt. 
Details across all domains in RAID  are reported in \uline{Appendix} \ref{appx:stats}, Table~\ref{tab-app:text_stats}. 
In addition, the outcomes of our statistical significance analysis of the perplexity-features computed for the RAID data   (\uline{Appendix} \ref{appx:stats}, Table~\ref{tab-app:feat_sig}) 
 confirm the initial   hypothesis of statistical  difference based on shuffling-induced perplexity-features between \hgt and \mgt.

\subsection{E1 --- Multi-Domain Detection}
\input{domain_results/domain_with_baselines}

Table~\ref{tab:domain-with-baselines} shows   results averaged over the 11 generators in RAID, for each of the domains.

First, \myalgo achieves a near-zero FPR ($0.001 \pm 0.001$) across all domains, making it $\sim$  
17x better than Fast-DetectGPT and 5x better than Binoculars, the two best competitors. 
 
Regarding FNR, \myalgo achieves the best scores in four out of eight domains (Poetry, Reddit, Reviews, Wiki), outperforming both competing methods by a substantial margin. 
For the remaining domains, while in Recipes and News the performance gap from the best method is only 0.001 and 0.04,  respectively, 
 the Abstracts domain requires closer inspection. In this regard,   we notice that most errors are associated with a single model family (Cohere). More generally, we investigated which  characteristics distinguish Abstracts texts from those of other domains: based on the text statistics of both fitting and test data (cf. Tables \ref{tab-app:text-gpt3_stats0}--\ref{tab-app:text-test_stats0}, \uline{Appendix}~\ref{appx:text-statistics}),    Abstracts is the only domain to have \hgt with very low   Flesch readability ease (FRE) scores, which range within 20 and 25, despite Abstracts being relatively shorter than texts in the other domains. 
The low FRE  indicates that  graduate-level education is  typically required  to   understand the text, as is common for  technical manuals in specialized fields, or academic research papers---and hence, Abstracts.  Note that, Abstracts \mgt have also lowest FRE. While this is comparably low to \hgt FRE on the test data (averaged across all  LLM generators in RAID),  the  fitting data (generated by GPT-3) show a relatively higher FRE (37), which might explain the still reasonably good performance of \myalgo.

\subsection{E2 --- Robustness to Adversarial Attacks}
\input{attacks_results/attacks_per_attack_rotated}
\input{multitude_results/multitude_per_lang_rotated}

Table~\ref{tab:attacks-per-attack} summarizes results by  \myalgo, and the best competitors from E1, 
on the RAID data under multiple types of adversarial attacks, averaged over the domains;  a per-domain view of these results is   available in \uline{Appendix}~\ref{appx:additional-res}, Table~\ref{tab:attacks-per-domain}.

Overall, \myalgo achieves a near-zero FPR across all 11 attacks, 
confirming its strong robustness. 
 In particular, it appears largely  
 immune to surface-level perturbations, such as character- and token-level manipulations.
For instance, the \textit{homoglyph} attack 
only increases \myalgo's FNR to $0.006$. In contrast, both Binoculars and Fast-DetectGPT degrade to near-complete failure, with $\text{FNR} = 0.979$ for both.  
A similar pattern is observed for the \textit{zero-width space} attack, which injects invisible characters that interfere with tokenization. 
More generally, perturbations such as number substitution, alternative spellings, and misspellings are handled with comparable effectiveness. This robustness likely stems from the fact that character-level noise is mitigated by the 
shuffling mechanism underlying \myalgo.

The most challenging attacks for the methods are transformations that somehow preserve discourse semantics.    \textit{Synonym substitution} yields the highest FNR for \myalgo ($0.316$), although it still substantially outperforms Binoculars ($0.703$) and Fast-DetectGPT ($0.734$). 
\textit{Paraphrasing} is also found challenging, as expected, since 
it can reduce the distributional gap between original and shuffled text, weakening the detection signal.
 \textit{Case perturbations} impacts  
the case-sensitivity of a byte-level BPE tokenizer like in GPT-Neo, thus affecting    
the resulting token sequences and hence  
the model’s probability estimates.
Notably, Fast-DetectGPT—also based on GPT-Neo—exhibits a similar, slightly worse degradation in FNR, supporting this hypothesis, while Binoculars relies on larger, more effective models (e.g., Falcon). 
Also, structural modifications such as article deletion, whitespace changes, and paragraph insertion 
do not significantly interfere with the shuffling-based signal exploited by \myalgo.

Note that FPR is not shown as the attacks only affect \mgt,  
so all detectors have a constant FPR across attacks---0.0 for \myalgo, 0.01 for Binoculars and 0.02 for Fast-DetectGPT.

 \vspace{-0.3mm}
\subsection{E3 --- Multilingual Detection}
 \vspace{-0.1mm}
Table~\ref{tab:multitude-per-lang} summarizes the detection results on the MULTITuDE benchmark across 18 languages (with abbreviations  described in \uline{Appendix}~\ref{appx:data}).   

\myalgo achieves a   FPR of $0.000 \pm 0.001$ and a  FNR of $0.087 \pm 0.128$   on average over the 18 languages.  
These results are even more notable given their substantial improvement over competitors, which largely collapse  in the multilingual scenario,  particularly for   non-Latin scripts (e.g., Chinese, Arabic, Cyrillic, Greek) and for Latin-based languages with rich diacritics (e.g., Czech, Polish).  
This is   not surprising, and also in line with findings from a recent study~\cite{LaCava2026maa}, since both Binoculars and Fast-DetectGPT suffer from the fact that the (byte-level BPE)  tokenization becomes more fragmented and probability estimates less stable under small perturbations, and hence their detection ability results heavily affected for those language families.  
By contrast, while sharing the same tokenizer, our \myalgo is much more robust since it exploits variations on the perplexities of a text and its shuffled version, with  the shuffling being substantially invariant  regardless of the language's morphological aspects.

\subsection{Impact of Proxy Model}
\label{sect:proxy} 
We argue that our perplexity-under-shuffling hypothesis reflects an intrinsic property of autoregressive generation. However, we investigate whether a different proxy model  produces variations in the magnitude of the perplexity signal, potentially leading to shifts in the resulting decision boundaries. 
To this purpose, we experimented with GPT2, which has  widely been used in detection contexts \cite{He2024mgtbench}. 
Comparing GPT-2 to GPT-Neo as a perplexity model, it differs mainly in training data (smaller and less diverse), scale (smaller), context length (shorter), and learned probability distributions (more English-centric and narrower, i.e., poorly  heterogeneous across domains and styles). 

We replicated all experiments  using GPT-2. \textit{Results are reported in \uline{Appendix}~\ref{appx:proxy}.}  
In summary, while we observe varying performance differences compared to the GPT-Neo-based results, two main remarks emerge: 
(i) Using GPT-2 tends to degrade the performance of \myalgo on task E2 (i.e., texts under attack) and, in particular, on task E3 (i.e., multilingual texts). This is not surprising, 
since GPT-2 was exposed to fewer stylistic variations during training, 
which may reduce its robustness to adversarial inputs, 
and it is heavily English-centric, leading to weaker and less stable probability estimates for non-English languages.
(ii) Regardless of the proxy model, the relative ranking of \myalgo w.r.t. the main competitors  remains unchanged for tasks E1, E2, and E3 (in terms of FNR); however, a slight yet expected increase in FPR (+0.061) suggests that GPT-Neo remains the preferable proxy model overall.

  \vspace{-1mm}
\section{Conclusion} 
 \vspace{-2mm}
We presented \myalgo, a training-free, model-agnostic statistical approach for \mgt detection that exploits a simple yet effective signal: the differential impact of text-shuffling on perplexity-features between \mgt and \hgt.
 Our evaluation of \myalgo under multiple real scenarios (8   domains, 11   attack types, 18 languages) has demonstrated state-of-the-art detection performance,  while being    cheaper than other black-box perturbation-based methods like  Binoculars  and  Fast-DetectGPT.  
Our ongoing work focuses on developing methods to tackle the detection of extremely short texts---an open challenge for  all existing approaches.

\section*{Limitations} 

\vspace{1.5mm}
\noindent\textbf{Extremely-short Text.\ }
Very short inputs (e.g., brief sentences) represent a pressing challenge for all current state-of-the-art \mgt detectors. Despite \myalgo can handle short text well, as noted in the remarkable results achieved on Reddit, addressing this shared limitation, e.g., by focusing on new shuffling strategies, remains a natural direction for further improvement.

\vspace{1.5mm}
\noindent\textbf{Robustness to Certain Attacks.\ }
Some of the attacks that provide semantics-preserving transformations 
pose  challenges to the detectors including \myalgo, as they can   attenuate the perplexity-under-shuffling signal. Even if \myalgo behaves as good as or even better than close competitors, 
further improving robustness to these attacks remains an open challenge.

\vspace{1.5mm}
\noindent\textbf{Combined Multilingual and Adversarial Scenarios.\ }
While our evaluation of \myalgo has shown its robustness separately under multilingual/cross-lingual settings and under the 11 adversarial attacks of the RAID benchmark, the latter are limited to English-only texts.

\section*{Ethical Considerations}
\noindent\textbf{Detectability of \mgt Content.\ }
Our findings highlight the inherent difficulty, faced by all state-of-the-art approaches, in reliably detecting \mgt under some conditions (e.g., attacks, languages). This means that sufficiently sophisticated text manipulation techniques over \mgt might evade detection, with potentially significant impact on individuals, communities, or society at large. Furthermore, no detector is  always immune to false positives, thus incorrectly flagging humans as machines. In light of these considerations, we strongly urge all parties involved to exercise caution and responsibility to ensure the safe and ethical deployment and utilization of these technologies.

\vspace{1.5mm}
\noindent\textbf{Broader Impact.\ }
The main goal of our research is to advance \mgt detection by proposing a method that is robust, efficient, and agnostic to the  text-generation model. We acknowledge that, during its systematic validation, we might have unveiled exploitable blind spots in current state-of-the-art detectors, including our own, which could be exploited for malicious purposes. We discard any responsibility for misuse of these findings and stress the importance of responsible and ethical use of these technologies by all actors involved.

\vspace{1.5mm}
\noindent\textbf{Transparency and Reproducibility.\ }
We are committed to publicly releasing all resources associated with this work, including the source code of \myalgo, the fitted distribution repository $\Omega$ across all settings, and an inference script, aiming at allowing the scientific community to validate and build upon our work, as well as fostering a broad adoption of \myalgo.

\appendix

\section{Additional methodological details on \myalgo}
\label{appx:details}

\paragraph{Uncertainty threshold.}

The \myalgo's decision logic is majority consensus over the perplexity-feature-specific   votes, i.e., $v_\phi = \mgt$ if $\hat{p}^{(\mgt)}_{z_\phi} \geq \hat{p}^{(\hgt)}_{z_\phi} - \Upsilon$,   $v_\phi = \hgt$ otherwise, with  $\Upsilon = 0$ by default. 
However, an alternative strategy that makes the individual feature decision more robust is to set $\Upsilon$ as a function  of $\hat{p}^{(\mgt)}_{z_\phi}$ and $\hat{p}^{(\hgt)}$.

To this purpose, we experimented with the following definitions of $\Upsilon(\hat{p}^{(\hgt)}_{z_\phi}, \hat{p}^{(\mgt)}_{z_\phi})$: 

 \begin{itemize}[itemsep=-2pt, leftmargin=*]
 \item $\hat{p}^{(\hgt)}_{z_\phi} - \hat{p}^{(\mgt)}_{z_\phi} + \epsilon$ \ (denoted as $\Upsilon_d$),
 \item 
$(\hat{p}^{(\hgt)}_{z_\phi} / \hat{p}^{(\mgt)}_{z_\phi}) -1 + \epsilon$ \ (denoted as $\Upsilon_r$ ),
\item 
$\log(\hat{p}^{(\hgt)}_{z_\phi} /\hat{p}^{(\mgt)}_{z_\phi}) + \epsilon$ \ (denoted as $\Upsilon_{lr}$),
\end{itemize}

\noindent
where $\epsilon$ is a small quantity set to 0.001, 0.025, and 0.05, respectively.

\paragraph{Outlier removal.}
We   acknowledge that the observed data, from any domain, might contain  outliers, i.e., values that inflate the standard deviation. This would happen due to the presence of  perplexity  values  computed for, e.g., very short texts,  or tokenization anomalies.

A common and statistically robust way to remove such outliers  is to compute the Inter-Quartile Range (IQR) over a set of measurements, i.e., a      value     is   considered an outlier if it lies outside $[Q_1-1.5\,\,IQR, Q_3+1.5\,\, IQR]$, where $Q_1$ and $Q_3$ denote the  first and third quartile of $z_\phi$, respectively, and $IQR = Q_3-Q_1$. 
 
The IQR  can hence be exploited as a 
filtering step at raw data level, which might  be applied before the  perplexity feature extraction.  
On RAID fitting-data, outlier removal affected an average of $5\% \pm 0.03$ (\hgt, \mgt) pairs across domains, which mainly regarded the cases mentioned above. 
In addition, and limited only to  older/smaller text-generation models (e.g.,  GPT2, MPT variants), we also found small samples of degenerated texts or very noisy texts that did not pass the plausibility check.

\section{Remarks on Computational Cost}
\label{appx:costs}
Let   $T$ denote the input length in tokens, $V$ the tokenizer vocabulary size, and   $C_\mathcal{M}$   the cost of a single forward pass of a decoder LLM  $\mathcal{M}$, with context window $W$. 
Methods like Binoculars and Fast-DetectGPT require two forward passes and, crucially, a full vocabulary aggregation of next-token probabilities for each position, due to the cross-perplexity and curvature numerator of the two methods, respectively. 
Their computational cost is hence $\mathcal{O}(T\, C_\mathcal{M}(W) +T\,V)$, where the second term accounts for the aforementioned aggregation step. 
By contrast, \myalgo only requires two inference passes while restricting perplexity computation to the observed token space only, with no access to the full vocabulary $V$. Therefore, its cost is $\mathcal{O}(T\,C_\mathcal{M}(W))$. This removes the $T\,V$ cost term, which is particularly impactful for modern LLMs with very large $V$. Note that, in practice, \myalgo also benefits from using a 3x smaller $\mathcal{M}$ than competing methods (2.7B vs. 7B), further reducing inference time and memory footprints.

\section{Details on benchmarks}
\label{appx:data}

\noindent\textbf{RAID}~\cite{dugan-etal-2024-raid}  
 provides \mgt (coupled with corresponding \hgt instances)  produced by 
11 LLM generators, namely \textit{ChatGPT, GPT-4, GPT-3, GPT-2, LLaMA-Chat, Mistral, Mistral-Chat, MPT, MPT-Chat, Cohere,} and \textit{Cohere-Chat}. 

Additionally, RAID provides  \mgt instances that underwent 
11 types of post-hoc adversarial attacks: 

 \begin{itemize}[itemsep=-2pt, leftmargin=*]
 \item 
\textit{Homoglyph}, which  replaces characters with visually-identical Unicode ones (e.g., e → e (U+0435)); 
\item \textit{Number substitution}, which  randomly shuffles digits of numbers; 
\item \textit{Article deletion}, which  removes articles (\textit{a}, \textit{an}, \textit{the}); \item \textit{Paragraph insertion}, which adds \textbackslash n\textbackslash n between sentences; 
\item \textit{Perplexity misspelling}, which inserts misspellings; 
\item \textit{Upper/lower}, which  swaps the case of words; \item \textit{Whitespace}, which adds spaces between characters;  \item \textit{Synonym substitution}, which swaps tokens with highly similar ones; \item \textit{Paraphrase}, which performs paraphrasing via a fine-tuned T5 model; \textit{Alternative spelling} adopts British English.
 \end{itemize}

\vspace{1.5mm}
\noindent \textbf{MULTITuDE} (v3)~\cite{macko-etal-2023-multitude,macko_2025_15519413},  
a multilingual benchmark of \mgt and \hgt news articles  in 18 languages.   
 
Specifically, it covers eight language families, namely
Indo-European---organized into \textit{Germanic}, including Dutch (nl), English (en), and German (de), 
\textit{Romanic}, including   Portuguese (pt), Romanian (ro), and Spanish (es), 
\textit{Slavic-Latin}, including Croatian (hr), Czech (cs), Polish (pl), Slovak (sk), and Slovenian (sl), 
\textit{Slavic-Cyrillic}, including Bulgarian (bg), Russian (ru), and Ukrainian (uk), and 
\textit{Hellenic}, represented by Greek (el)--- 
\textit{Uralic}, represented by Hungarian (hu), 
\textit{Semitic}, represented by  Arabic (ar), 
and \textit{Sino-Tibetan}, represented by Chinese (zh).  This organization also
corresponds to five writing scripts (12  Latin,
3$\times$Cyrillic, 1$\times$Arabic, 1$\times$Hanzi, and 1$\times$Greek). 

MULTITuDE contains human-written news articles and machine-generated counterparts, which are produced by seven LLMs prompted with the original headlines of the articles. 
For our evaluation, we used the test subset of 5395 \mgt instances, and associated \hgt instances,   corresponding to the most recent model among those available in MULTITuDE, i.e.,  GPT3.5-turbo.

\section{Details on competing detectors}
\label{app:competing detectors}

\noindent \textbf{Log-Likelihood}~\cite{solaiman2019release}: this method scores a text according to the average token-wise log probability produced by a language model, where the higher the scores, the higher the likelihood of a text to be machine-generated.

\vspace{1.5mm}
\noindent \textbf{Rank}~\cite{gehrmann-etal-2019-gltr}: this method scores a text using the average rank of its words, with individual ranks for each word determined using the preceding context. Lower scores indicate a higher probability that the text is machine-generated.

\vspace{1.5mm}
\noindent \textbf{Log-Rank}~\cite{pmlr-v202-mitchell23a}: unlike Rank, this variant first applies a logarithmic transformation to each word rank before averaging.

\vspace{1.5mm}
\noindent  \textbf{Entropy}~\cite{gehrmann-etal-2019-gltr}:
similar to Rank, it averages the entropy value of each word conditioned on the preceding context, with lower entropy denoting higher \mgt likelihood.

\vspace{1.5mm}
\noindent \textbf{GLTR}~\cite{gehrmann-etal-2019-gltr}: this method extracts the fraction of words that rank within a certain position (e.g., 10, 100, 1,000) in a given text, thus serving as a tool for the subsequent classification tasks.

\vspace{1.5mm}
\noindent  \textbf{LRR}~\cite{su-etal-2023-detectllm}:
it combines the aforementioned Log-Likelihood and Log-Rank scores.

All the above methods are provided through  MGTBench~\cite{He2024mgtbench} at 
\url{https://github.com/TrustAIRLab/MGTBench/}.

\vspace{1.5mm}
\noindent  \textbf{Fast-DetectGPT}~\cite{bao2024fastdetectgpt}: this method determines whether a text is machine-generated using word choice discrepancies between LLMs and humans, obtained via conditional probability curvature. 

\vspace{1.5mm}
\noindent  \textbf{Binoculars}~\cite{hans2024binoculars}: this measures cross-model likelihood ratio between a scorer and a reference model, scoring how anomalously one model rates the text relative to another as an indicator of whether a text has been machine-generated.

\section{Details on experimental settings of \myalgo}
\label{appx:settings}

 \paragraph{Distribution families and perplexity features.}
For the bootstrapping in our approach, we considered a set of continuous probability distribution families\footnote{Implemented via the Python \texttt{scipy.stats} module.} in order to capture a wide range of possible data-generating behaviors, spanning light- and heavy-tailed distributions, symmetric and asymmetric forms, as well as bounded and unbounded supports, thereby providing a comprehensive basis for robust distributional fitting: 
\textit{Normal, LogNormal,  Student's t, Exponential, Powerlaw, Gamma, Weibull, Beta, Burr, Pareto, Generalized Extreme Value}. 

However, throughout all evaluation tasks, we found that Gamma  (for any perplexity-feature type) and  Burr (for \textit{sum} and \textit{ratio} perplexity-feature types) are consistently the best-fitting distribution models across all RAID domains; see   Table \ref{tab:configs} for details on the best-performing settings of \myalgo.

We also found that, across all domains, using  \textit{change} as a predictor (i.e., as a member of $\mathcal{E}$) was redundant in most cases, which is not surprising given its close relation  to \textit{ratio} and \textit{log-ratio}. The latter, instead, were found to be used (often interchangeably) along with \textit{sum} and \textit{diff}  as the most effective  predictors.

 \textit{The fitting-distribution repositories will be made publicly available upon acceptance of this work. }

\paragraph{Fitting and test data.}
All \myalgo results presented throughout this work corresponded to the application at inference  of the fitting-distribution repositories produced for each domain-specific  80\% split of the RAID subset of \mgt generated by GPT-3. 
For the tasks E1 and E2, the test data for both \myalgo and competing  methods corresponded to the 20\% splits of the RAID data, with \mgt from all generators, for each domain.
Moreover, for task E3, \myalgo used the distribution repository associated with the  RAID news domain, where \mgt were selected from the subset of GPT-3.  
Note that we chose  GPT-3 when fitting the perplexity-feature distributions as it serves as a prototypical representative of  most modern LLMs, and is empirically shown to generalize well; for instance, the near-zero FPR achieved across all 11 RAID generators (Table~\ref{tab:domain-with-baselines}) confirms that this reference model effectively covers the \mgt feature space.

\input{appendix_material/configs}

\paragraph{Selecting the best configurations.}
Table \ref{tab:configs} summarizes the best-performing settings of \myalgo, for each evaluation task and for each modality of criterion optimization (i.e., lowest FNR, lowest FPR).  
These configurations were determined via grid-search over all combinations of three components: the outlier filtering strategy, the implausibility check, and the uncertainty threshold function.  
For each evaluation task and modality, the best configuration was determined by majority voting across generators (11 for E1), domain-attack combinations (8x11 for E2), or languages (18 for E3), with ties resolved by selecting the configuration that yields the highest mean target rate.

Two consistent patterns emerge from Table~\ref{tab:configs}. 
In the low-FPR mode, the implausibility check is enabled in all configurations, and the outlier filtering is enabled in the majority of cases. This reflects the conservative low-FPR perspective: by discarding evaluations that are out-of-distribution for both \mgt and \hgt fitted densities, they reduce false-positive detections on \hgt.
Conversely, in the low-FNR mode, less 
conservative filtering is needed for  
maximizing sensitivity to \mgt.

\section{Text statistics on RAID data}
\label{appx:text-statistics}

Tables \ref{tab-app:text-gpt3_stats0}--\ref{tab-app:text-test_stats0} provide a summary of the  main textual  statistics of \mgt and \hgt RAID data. 
Specifically, the tables report the total number of instances (documents) for each domain; 
the average (per-document) number of syllables, words, and paragraphs; 
the average number of sentences on average per-paragraph and per-document; 
the average \textit{Flesch Reading Ease} as a readability measure,\footnote{We used the implementation available from the Python \texttt{textstat} package} and the average  \textit{Compression Ratio}, which is computed as the ratio between the original size of a text and its compressed (gzip) size. 

Note that the Flesch Reading Ease   is calculated as a linear combination of the   average number of words per sentence and the average number of syllables per word. Texts with smaller number of  words and sentences have   higher  score, i.e., are   easier to understand:
$$
{\small
206.835-1.015\left({\frac {\text{\# words}}{\text{\# sentences}}}\right)-84.6\left({\frac {\text{\# syllables}}{\text{\# words}}}\right).
}
$$

\section{Statistics on  perplexities and perplexity-features of RAID data}
\label{appx:stats}

Table~\ref{tab-app:text_stats} shows  basic statistics of \mgt and \hgt perplexities and perplexity-features in RAID data, over all domains. 
While the central tendency and variability of perplexity is much larger on \hgt than \mgt, 
perplexity on texts after shuffling  tends to be higher on \mgt than  \hgt.  
More comparative evidence is provided by the perplexity-feature statistics. Indeed, the difference between perplexity on shuffled text and perplexity on original text  is   significantly higher on \mgt than \hgt. 
Also, considering the ratios, perplexity on shuffled  \mgt is typically 4 to 5x larger than perplexity on the original text (systematic multiplicative effect), while for \hgt is only $\sim$20–60\% larger than perplexity on the original text, for most cases. 
Yet, relative changes are extremely large for \mgt, which indicates massive proportional increases driven by small perplexity on the original text, while relative changes for \hgt are still large, but far smaller than for \mgt.

 Table~\ref{tab-app:text_stats_gpt2ppl} reports analogous statistics using GPT-2 as the perplexity model. 
 Compared to the GPT-Neo-based setting, we observe a substantial overall increase in the perplexity gap between \mgt and \hgt, along with a moderate to small increase in the perplexity gap between \mgt and shuffled \hgt texts, as well as in the corresponding features.  

Table~\ref{tab-app:feat_sig} summarizes the statistical significance  of the perplexity-features computed for the RAID \hgt and \mgt instances. 
Effect sizes, confidence intervals, and p-values are reported for the Welch’s t-tests and the Mann-Whitney U tests applied to the feature-specific sample pairs of \hgt and \mgt for each RAID domain (GPT-3 is selected as \mgt model). 
It can be noted that, in the vast majority of cases, p-values are extremely low; therefore, the null hypothesis can be rejected. 
In addition, practically meaningful effects are observed with respect to all features across all domains, with some exceptions in Abstracts, Reddit and Reviews. 
As shown in Table~\ref{tab-app:feat_sig_gpt2ppl}, using GPT-2 as the perplexity model leads to an even stronger tendency to reject the null hypothesis, with only three exceptions (in Reddit and Reviews) out of the forty total (domain, feature-type) combinations.

\section{Additional performance results}
\label{appx:additional-res}

\input{attacks_results/attacks_per_domain}

\subsection{Insights into the per-domain \myalgo results on task E2}
Table \ref{tab:attacks-per-domain} complements Table~\ref{tab:attacks-per-attack} by providing a per-domain aggregation of the results  
achieved by \myalgo and competing methods across the RAID texts that underwent various attacks.

Notably, \myalgo maintains near-zero FPR across all domains (at most $0.002$ for Reviews), confirming that its strong robustness to attacks is not domain-sensitive.

Concerning FNR, \myalgo achieves near-zero values ($0.000$--$0.003$) on 5 out of 8 domains, outperforming both Binoculars ($0.155$--$0.267$) and Fast-DetectGPT ($0.227$--$0.453$) by a wide margin.  These domains (i.e., Reddit, Reviews, Wiki, and Poetry, Recipes)  exhibit stylistically distinctive discourse structures, where sentence shuffling produces a   signal  suitably exploited by \myalgo. 
Despite an increase in FNR, the advantage is also preserved on Books, with \myalgo exhibiting half the FNR of Binoculars, and one third the FNR of Fast-DetectGPT. 
Abstracts and News under attacks appear to be  more challenging for \myalgo, 
where Binoculars exhibits lower FNR in both domains, and Fast-DetectGPT only for News. 

Overall, averaging across all 8 domains, \myalgo achieves lower FPR and FNR than competing approaches, thus confirming its domain-independent robustness to adversarial attacks.

\subsection{Details on sensitivity  to proxy model}
\label{appx:proxy}

Table \ref{tab:proxy-sensitivity} summarizes results achieved by \myalgo across all evaluation tasks, by highlighting the comparison between the use of GPT-Neo and GPT-2 as perplexity model. (FPR for E2 is not reported as attacks only involve \mgt.)
See main text (Sect. \ref{sect:proxy}) for  discussion on these results.

\begin{table}[t!]
\centering
  \setlength{\tabcolsep}{2pt}
  \renewcommand{\arraystretch}{1.15}
  \scalebox{0.8}{
\begin{tabular}{ll|ccc}
\toprule
\multirow{2}{*}{Task} & \multirow{2}{*}{Evaluation}
  & \multicolumn{3}{c}{FPR} \\
\cmidrule(lr){3-5}
  &
  & GPT-Neo & GPT-2 & $\Delta$ \\
\midrule
E1 & RAID domains
  & $.001 \pm .001$ & $.001 \pm .001$ & \textit{.000} \\

E2 & Adversarial attacks
  & -- & -- & -- \\

E3 & MULTITuDE langs.
  & $.000 \pm .001$ & $.061 \pm .052$ & \textit{+.061} \\
\bottomrule

\toprule
\multirow{2}{*}{Task} & \multirow{2}{*}{Evaluation}
  & \multicolumn{3}{c}{FNR} \\
\cmidrule(lr){3-5}
  &
  & GPT-Neo & GPT-2 & $\Delta$ \\
\midrule
E1 & RAID domains
  & $.031 \pm .054$ & $.025 \pm .040$ & \textit{-.006} \\

E2 & Adversarial attacks
  & $.094 \pm .097$ & $.123 \pm .123$ & \textit{+.029} \\

E3 & MULTITuDE langs.
  & $.087 \pm .128$ & $.137 \pm .256$ & \textit{+.050} \\
\bottomrule
\end{tabular}
}
  \caption{Comparison of \myalgo performance errors across all tasks by varying perplexity model.}
  \label{tab:proxy-sensitivity}
\end{table}

%================================== 
% STATS TABLES
%===============================

\input{appendix_material/text_stats}

\input{appendix_material/feature_sig}

\end{document}

%% file: domain_results/domain_with_baselines.tex
\begin{table*}[t!]
  \centering
  \setlength{\tabcolsep}{3pt}
  \renewcommand{\arraystretch}{1.15}
  \scalebox{0.79}{
  \begin{tabular}{l|cccccccccccccccc|cc}
    \toprule
    Domain & \multicolumn{2}{c}{\textbf{Abstracts}} & \multicolumn{2}{c}{\textbf{Books}} & \multicolumn{2}{c}{\textbf{News}} & \multicolumn{2}{c}{\textbf{Poetry}} & \multicolumn{2}{c}{\textbf{Recipes}} & \multicolumn{2}{c}{\textbf{Reddit}} & \multicolumn{2}{c}{\textbf{Reviews}} & \multicolumn{2}{c}{\textbf{Wiki}} & \multicolumn{2}{|c}{mean $\pm$ std} \\
    \cmidrule(lr){2-3} \cmidrule(lr){4-5} \cmidrule(lr){6-7} \cmidrule(lr){8-9} \cmidrule(lr){10-11} \cmidrule(lr){12-13} \cmidrule(lr){14-15} \cmidrule(lr){16-17} \cmidrule(l){18-19}
    Detector & \scriptsize FPR & \scriptsize FNR & \scriptsize FPR & \scriptsize FNR & \scriptsize FPR & \scriptsize FNR & \scriptsize FPR & \scriptsize FNR & \scriptsize FPR & \scriptsize FNR & \scriptsize FPR & \scriptsize FNR & \scriptsize FPR & \scriptsize FNR & \scriptsize FPR & \scriptsize FNR & \scriptsize FPR & \scriptsize FNR \\
    \midrule
    Ours   & \textbf{.002} & .153 & \textbf{.000} & .015 & \textbf{.000} & .064 & \textbf{.000} & \textbf{.013} & \textbf{.000} & .003 & \textbf{.000} & \textbf{.000} & \textbf{.002} & \textbf{.000} & .001 & \textbf{.000} & \textbf{.001 $\pm$ .001} & .031 $\pm$ .054 \\

    \midrule
    Binoculars & .008 & \textbf{.021} & .003 & \textbf{.005} & \textbf{.000} & \textbf{.021} & .003 & .025 & \textbf{.000} & \textbf{.002} & .008 & .037 & .021 & .011 & \textbf{.000} & .011 & .005 $\pm$ .007 & \textbf{.017 $\pm$ .012} \\
    FastDetect & .008 & .033 & .017 & .009 & .017 & .023 & .025 & .134 & .006 & .023 & .020 & .104 & .021 & .013 & .020 & .015 & .017 $\pm$ .007 & .044 $\pm$ .047 \\
    \midrule
    Entropy & .216 & .203 & .164 & .153 & .225 & .201 & .167 & .171 & .107 & .079 & .174 & .141 & .212 & .183 & .194 & .232 & .182 $\pm$ .038 & .170 $\pm$ .047 \\
    GLTR & .072 & .086 & .013 & .032 & .029 & .040 & .107 & .088 & .070 & .022 & .072 & .061 & .040 & .036 & .032 & .040 & .054 $\pm$ .031 & .051 $\pm$ .025 \\
    Log-Like & .063 & .066 & .018 & .027 & .024 & .031 & .119 & .108 & .044 & .030 & .061 & .060 & .030 & .032 & .029 & .042 & .048 $\pm$ .033 & .050 $\pm$ .028 \\
    Rank & .263 & .137 & .101 & .081 & .226 & .144 & .235 & .112 & .291 & .179 & .227 & .116 & .156 & .109 & .165 & .113 & .208 $\pm$ .063 & .124 $\pm$ .029 \\
    Log-Rank & .070 & .071 & .015 & .026 & .027 & .037 & .104 & .086 & .039 & .022 & .064 & .063 & .029 & .032 & .024 & .036 & .046 $\pm$ .030 & .047 $\pm$ .023 \\
    LRR & .083 & .140 & .011 & .053 & .037 & .081 & .069 & .121 & .033 & .048 & .058 & .103 & .036 & .060 & .018 & .062 & .043 $\pm$ .025 & .084 $\pm$ .034 \\
    \bottomrule
  \end{tabular}}
  \caption{(E1)  Errors  of \myalgo and competing methods   on RAID test data. Per-domain  FPR and FNR scores are averaged over multiple (11) generators.  Rightmost columns report mean $\pm$ std across domains.}
  \label{tab:domain-with-baselines}
  \vspace{-3mm}
\end{table*}

%% file: attacks_results/attacks_per_attack_rotated.tex
\begin{table*}[t!]
  \centering
  \setlength{\tabcolsep}{3pt}
  \renewcommand{\arraystretch}{1}
  \resizebox{\textwidth}{!}{
  \begin{tabular}{l|ccccccccccc|c}
    \toprule
    Detector
      & {\textbf{Homoglyph}}
      & {\textbf{Number}}
      & {\textbf{Art. del.}}
      & {\textbf{Ins.\ paragr.}}
      & {\textbf{Misspell.}}
      & {\textbf{Up/low}}
      & {\textbf{Whitesp.}}
      & {\textbf{0-width sp.}}
      & {\textbf{Synonym}}
      & {\textbf{Paraphr.}}
      & {\textbf{Alt.\ spell.}}
      & mean $\pm$ std\\
    \midrule
    Ours        & \textbf{.006} & .018 & .101 & .068 & .028 & .160 & .121 & \textbf{.002} & \textbf{.316} & .187 & .028 & \textbf{.094 $\pm$ .097} \\

    Binoculars  & .979 & \textbf{.008} & \textbf{.011} & \textbf{.012} & \textbf{.007} & \textbf{.058} & \textbf{.087} & .408 & .703 & \textbf{.138} & \textbf{.006} & .220 $\pm$ .334 \\
    FastDetect  & .979 & .107 & .099 & .056 & .050 & .175 & .215 & .966 & .734 & .171 & .041 & .327 $\pm$ .373 \\
    \bottomrule
  \end{tabular}}
  \caption{(E2) FNR of \myalgo and competing methods under 11 adversarial text attacks, averaged over the 8 domains. Per-attack best (lowest) FNR are in   boldface. Rightmost column reports mean $\pm$ std across attacks.}
  \label{tab:attacks-per-attack}
\end{table*}

%% file: multitude_results/multitude_per_lang_rotated.tex
\begin{table*}[t!]
  \centering
  \setlength{\tabcolsep}{3pt}
  \renewcommand{\arraystretch}{1.1}
  \resizebox{\textwidth}{!}{
  \begin{tabular}{cl|cccccccccccccccccc|c}
    \toprule
     &  Detector  & \textbf{ar} & \textbf{bg} & \textbf{cs} & \textbf{de} & \textbf{el} & \textbf{en} & \textbf{es} & \textbf{hr} & \textbf{hu} & \textbf{nl} & \textbf{pl} & \textbf{pt} & \textbf{ro} & \textbf{ru} & \textbf{sk} & \textbf{sl} & \textbf{uk} & \textbf{zh} &  mean $\pm$ std \\
    \midrule
    \multirow{3}{*}{\rotatebox[origin=c]{90}{\textbf{FPR}}}
      & Ours         & \textbf{.000} & \textbf{.000} & \textbf{.000} & \textbf{.000} & \textbf{.000} & \textbf{.000} & \textbf{.000} & \textbf{.000} & \textbf{.000} & \textbf{.000} & \textbf{.000} & \textbf{.000} & \textbf{.000} & \textbf{.000} & \textbf{.000} & .004 & \textbf{.000} & \textbf{.000} & \textbf{.000 $\pm$ .001} \\

      & Binoculars   & \textbf{.000} & \textbf{.000} & \textbf{.000} & .010 & \textbf{.000} & .007 & .013 & \textbf{.000} & \textbf{.000} & \textbf{.000} & \textbf{.000} & \textbf{.000} & \textbf{.000} & \textbf{.000} & \textbf{.000} & \textbf{.000} & \textbf{.000} & .013 & .002 $\pm$ .005 \\
      & FastDetect   & \textbf{.000} & .007 & \textbf{.000} & .007 & .010 & .013 & .010 & .010 & .007 & .007 & .007 & .023 & .007 & .023 & .017 & .010 & .007 & .007 & .009 $\pm$ .006 \\
    \midrule
    \multirow{3}{*}{\rotatebox[origin=c]{90}{\textbf{FNR}}}
      & Ours          & \textbf{.000} & \textbf{.000} & \textbf{.113} & \textbf{.027} & \textbf{.000} & \textbf{.001} & \textbf{.007} & \textbf{.170} & \textbf{.213} & \textbf{.100} & \textbf{.057} & \textbf{.010} & \textbf{.007} & \textbf{.000} & \textbf{.220} & \textbf{.513} & \textbf{.000} & \textbf{.129} & \textbf{.087 $\pm$ .128} \\

      & Binoculars   & 1.000 & .997 & 1.000 & .757 & .997 & .003 & .243 & .997 & 1.000 & .933 & .980 & .437 & .970 & 1.000 & 1.000 & 1.000 & 1.000 & .508 & .823 $\pm$ .310 \\
      & FastDetect   & .887 & .680 & .733 & .460 & .857 & .017 & .283 & .630 & .683 & .550 & .693 & .367 & .240 & .383 & .820 & .813 & .653 & .681 & .580 $\pm$ .240 \\
    \bottomrule
  \end{tabular}}
  \caption{(E3) Errors of  \myalgo and competing methods on      MULTITuDE data, over various languages. 
  Per-language best (lowest) scores are in boldface. Rightmost column reports mean $\pm$ std across languages.}
  \label{tab:multitude-per-lang}
  \vspace{-3mm}
\end{table*}

%% file: appendix_material/configs.tex
\begin{table}[t!]
  \centering
  \setlength{\tabcolsep}{4pt}
  \renewcommand{\arraystretch}{1.15}
  \scalebox{0.82}{
  \begin{tabular}{ll ccc ccc}
    \toprule
    & & \multicolumn{3}{c}{\textbf{low-FPR config}}
      & \multicolumn{3}{c}{\textbf{low-FNR config}} \\
    \cmidrule(lr){3-5}\cmidrule(l){6-8}
    \textbf{Task} & \textbf{Domain}
      & \small Outl. & \small Impl.\ & \small Uncert.
      & \small Outl. &  \small Impl.\ & \small Uncert. \\
    \midrule
    \multirow{8}{*}{\textbf{E1}}
      & Abstracts
        & \xmark & \cmark & $\Upsilon_{d}$
        &  \cmark 
        & \xmark & $\Upsilon_{d}$ \\
      & Books
        & \cmark 
        & \cmark & $\Upsilon_{lr}$
        & \xmark   & \xmark & $\Upsilon_{d}$ \\
      & News
        & \cmark & \cmark & $\Upsilon_{d}$
        &   \cmark
        & \xmark & $\Upsilon_{d}$ \\
      & Poetry
        & \cmark  & \cmark & 0
        & \xmark   & \xmark & $\Upsilon_{d}$ \\
      & Recipes
        & \cmark  & \cmark & $\Upsilon_{d}$
        & \xmark   & \xmark & $\Upsilon_{d}$ \\
      & Reddit
        & \cmark  & \cmark & $\Upsilon_{lr}$
        & \xmark   & \cmark & $\Upsilon_{d}$ \\
      & Reviews
        & \cmark  & \cmark & $\Upsilon_{d}$
        &  \cmark 
        & \xmark & $\Upsilon_{d}$ \\
      & Wiki
        & \cmark  & \cmark & $\Upsilon_{d}$
        & \xmark   & \cmark & $\Upsilon_{d}$ \\
    \midrule
    \textbf{E2} & \textit{Any}
        & \cmark  & \cmark & 0
        & \xmark   & \xmark & $\Upsilon_{d}$ \\
    \midrule
    \textbf{E3}
      & \textit{Any} % fitted on news
        & \cmark  & \cmark & $\Upsilon_{lr}$
        & \xmark &   \xmark & $\Upsilon_{d}$ \\
    \bottomrule
  \end{tabular}}
  \caption{%
    Best-performing settings of \myalgo for each evaluation task. Abbreviation \textit{Outl.} stands for outlier removal filter (enabled or not),  
    \textit{Impl.} for implausibility check (enabled or not), and \textit{Uncert.} for the uncertainty threshold function setting.
  }
  \label{tab:configs}
\end{table}

%% file: attacks_results/attacks_per_domain.tex
\begin{table}[t!]
  \centering
  \setlength{\tabcolsep}{3pt}
  \renewcommand{\arraystretch}{1.15}
  \scalebox{0.85}{
  \begin{tabular}{lcccccc}
    \toprule
    \textbf{Detector} & \multicolumn{2}{c}{\textbf{Ours}} & \multicolumn{2}{c}{\textbf{Binoculars}} & \multicolumn{2}{c}{\textbf{FastDetect}} \\
    \cmidrule(lr){2-3} \cmidrule(lr){4-5} \cmidrule(l){6-7}
    \textbf{Domain} & \scriptsize FPR & \scriptsize FNR & \scriptsize FPR & \scriptsize FNR & \scriptsize FPR & \scriptsize FNR \\
    \midrule
    Abstracts & \textbf{.001} & .278 & .005 & \textbf{.197} & .007 & .318 \\
    Books & \textbf{.001} & \textbf{.100} & .003 & .187 & .028 & .244 \\
    News & \textbf{.000} & .371 & .001 & \textbf{.250} & .015 & .251 \\
    Poetry & \textbf{.000} & \textbf{.000} & .008 & .155 & .016 & .453 \\
    Recipes & \textbf{.000} & \textbf{.003} & .003 & .224 & .013 & .444 \\
    Reddit & \textbf{.000} & \textbf{.000} & .012 & .223 & .014 & .444 \\
    Reviews & \textbf{.002} & \textbf{.000} & .018 & .254 & .023 & .227 \\
    Wiki & \textbf{.001} & \textbf{.000} & .003 & .267 & .017 & .232 \\
    \midrule
    mean & \textbf{.000} & \textbf{.094} & .007 & .220 & .017 & .327 \\
    $\pm$ std & \textbf{.001} & \textbf{.148} & .006 & .038 & .006 & .103 \\
    \bottomrule
  \end{tabular}}
  \caption{
  (E2) Performance of \myalgo and competing methods on RAID data with attacks, by varying domain. Results are averages over all the 11 attacks. 
  Bottom row reports mean $\pm$ std across domains.}
   \label{tab:attacks-per-domain}
\end{table}

%% file: appendix_material/text_stats.tex
\begin{table*}[t] 
\centering
\scalebox{0.85}{
\setlength{\tabcolsep}{5pt}
\renewcommand{\arraystretch}{1.15}
\begin{tabular}{l|l|cccccccc} 
\toprule
 &  & Abstracts & Books & News & Poetry & Recipes & Reddit & Reviews & Wiki \\
 \midrule
 \multirow{8}{*}{\rotatebox{90}{HGT}} & \#docs &  1412 & 1424 & 1424 & 1416 & 1417 & 1423 & 754 & 1423 \\
 & per-doc \#syllables &  331.33 &    643.20 &    568.56 &    471.06 &    414.75 &    243.40 &    567.88 &    334.79 \\
 & per-doc \#words &   173.54 &    420.46 &    368.04 &    355.38 &    292.69 &    178.54 &    376.15 &    200.49 \\
 & per-doc \#paragraphs &   1.00 &    1.00 &    1.00 &    1.00 &    1.00 &    1.00 &    1.00 &    1.00 \\
 & per-doc, per-parag. \#sentences &   7.94 &    19.62 &    18.84 &    14.33 &    18.61 &    9.37 &    19.89 &    9.71 \\
 & per-doc 1-sen. \#paragraphs &   0.000 &    0.000 &    0.000 &    0.083 &    0.002 &    0.020 &    0.012 &    0.001 \\
 & per-doc 2-sen. \#paragraphs &  0.000 &    0.000 &    0.000 &    0.045 &    0.001 &    0.015 &    0.015 &    0.003 \\
 & per-doc  Flesch Reading Ease &  22.15 &    52.90 &    56.07 &    42.79 &    69.11 &    65.47 &    59.63 &    42.63 \\
 & per-doc Compression Ratio &  2.90 &    3.04 &    3.01 &    3.03 &    3.13 &    2.78 &    2.87 &    2.82 \\
\midrule
 \multirow{8}{*}{\rotatebox{90}{MGT}} & \#docs &  1412 & 1424 & 1424 & 1416 & 1417 & 1423 & 754 & 1423 \\
 & per-doc \#syllables &  173.50 &    194.99 &    278.19 &    115.61 &    210.10 &    138.27 &    166.74 &    299.74 \\
 & per-doc \#words &   98.61 &    133.05 &    184.93 &    93.66 &    144.14 &    102.28 &    110.07 &    186.29 \\
 & per-doc \#paragraphs &  1.08 &    2.94 &    9.60 &    10.26 &    12.81 &    2.96 &    2.38 &    5.06 \\
 & per-doc, per-parag. \#sentences &  4.78 &    3.07 &    1.13 &    1.15 &    2.10 &    3.16 &    3.32 &    2.51 \\
 & per-doc  Flesch Reading Ease &   37.04 &    64.39 &    60.94 &    43.03 &    67.57 &    77.57 &    62.18 &    53.14 \\
 & per-doc Compression Ratio &   2.77 &    2.54 &    2.89 &    2.96 &    2.79 &    2.52 &    2.37 &    3.03 \\
\bottomrule
\end{tabular}
}
\caption{Main text statistics of \mgt and \hgt  averaged over RAID   data, for each domain. \mgt statistics here refer to texts generated by GPT-3, which were used  for fitting. 
}
\label{tab-app:text-gpt3_stats0}
 \end{table*}

\begin{table*}[t] 
\centering
\scalebox{0.85}{
\setlength{\tabcolsep}{5pt}
\renewcommand{\arraystretch}{1.15}
\begin{tabular}{l|l|cccccccc} 
\toprule
 &  & Abstracts & Books & News & Poetry & Recipes & Reddit & Reviews & Wiki \\
 \midrule
 \multirow{8}{*}{\rotatebox{90}{HGT}} & \#docs & 3894 & 3927 & 3916 & 3905 & 3905 & 3916 & 2079  & 3916  \\  
 & per-doc \#syllables &  350.64 &    769.60 &    581.55 &    468.32 &    437.47 &    241.60 &    468.23 &    341.74 \\
 & per-doc \#words &  182.63 &    505.31 &    373.66 &    360.17 &    310.03 &    176.47 &    308.03 &    204.24 \\
 & per-doc \#paragraphs &  1.00 &    1.00 &    1.00 &    1.00 &    1.00 &    1.00 &    1.00 &    1.00 \\
 & per-doc, per-parag. \#sentences &  8.40 &    24.05 &    19.17 &    14.62 &    19.76 &    9.08 &    16.95 &    9.94 \\
 & per-doc Flesch Reading Ease &  21.10 & 53.37 & 55.35 & 48.02 & 70.56 & 64.20 & 58.90 & 42.69 \\
 & per-doc Compression Ratio & 2.92 & 3.13 & 3.03 & 3.04 & 3.17 & 2.71 & 2.76 & 2.83 \\
\midrule
 \multirow{8}{*}{\rotatebox{90}{MGT}} & \#docs & 3894 & 3927 & 3916 & 3905 & 3905 & 3916 & 2079  & 3916  \\  
 & per-doc \#syllables &  356.12 & 405.11 & 460.87 & 268.38 & 324.75 & 303.56 & 398.71 & 496.76 \\
 & per-doc \#words &  194.56 & 269.54 & 294.22 & 209.69 & 228.12 & 219.07 & 265.14 & 297.92 \\
 & per-doc \#paragraphs &  2.44 &    4.18 &    9.73 &    18.58 &    15.18 &    3.57 &    3.58 &    6.26 \\
 & per-doc, per-parag. \#sentences &  6.32 &    5.58 &    3.13 &    1.60 &    2.73 &    9.41 &    7.86 &    4.95 \\
 & per-doc  Flesch Reading Ease &  24.14 &    57.34 &    50.72 &    10.72 &    33.53 &    69.80 &    50.99 &    45.43 \\
 & per-doc Compression Ratio &  5.02 &    4.29 &    4.90 &    7.38 &    4.51 &    5.69 &    5.74 &    4.79 \\

\bottomrule
\end{tabular}
}
\caption{Main text statistics of \mgt and \hgt  averaged over RAID test data, for each domain. \mgt statistics refer to averages over all generators.  
}
\label{tab-app:text-test_stats0}
 \end{table*}

\clearpage
 \begin{sidewaystable*}[t!]
\centering
\scalebox{0.8}{
\setlength{\tabcolsep}{3pt}
\renewcommand{\arraystretch}{1.15}
\begin{tabular}{l|l|ccccc|ccccc|cc|cc|cc|cc} 
\toprule
Domain & Type & Mean & Median &	Stdev &Min &Max  &	Mean& Median & Stdev 	 & Min & Max & Mean & Median	& Mean & Median 	& Mean 	& Median  	& Mean& Median \\
& & 
$ppl$ & $ppl$ &	$ppl$ & $ppl$ &	$ppl$ & $ppl_{shuf}$ & $ppl_{shuf}$ &	$ppl_{shuf}$ & $ppl_{shuf}$ &	$ppl_{shuf}$ & diff & diff & ratio  & ratio & logratio & logratio & change & change \\
\midrule

\multirow{2}{*}{Abstracts}
 & MGT & 4.807 & 4.496 & 1.682 & 1.966 & 28.830 & 7.335 & 6.029 & 13.112 & 2.287 & 393.822 & 2.528 & 1.323 & 1.532 & 1.289 & 0.330 & 0.253 & 53.195 & 28.942 \\
 &  HGT & \cellcolor{gray!20}11.726 & \cellcolor{gray!20}10.949 & \cellcolor{gray!20}4.310 & \cellcolor{gray!20}2.433 & \cellcolor{gray!20}48.305 & \cellcolor{gray!20}15.510 & \cellcolor{gray!20}14.789 & \cellcolor{gray!20}5.427 & \cellcolor{gray!20}2.486 & \cellcolor{gray!20}50.731 & \cellcolor{gray!20}3.784 & \cellcolor{gray!20}3.391 & \cellcolor{gray!20}1.336 & \cellcolor{gray!20}1.301 & \cellcolor{gray!20}0.278 & \cellcolor{gray!20}0.263 & \cellcolor{gray!20}33.562 & \cellcolor{gray!20}30.073 \\

\multirow{2}{*}{Books}
 & MGT & 5.416 & 5.190 & 1.315 & 2.609 & 16.760 & 8.630 & 7.140 & 12.326 & 3.451 & 353.140 & 3.214 & 1.749 & 1.635 & 1.331 & 0.370 & 0.284 & 63.516 & 33.115 \\
 &  HGT & \cellcolor{gray!20}19.336 & \cellcolor{gray!20}18.262 & \cellcolor{gray!20}6.300 & \cellcolor{gray!20}5.601 & \cellcolor{gray!20}100.578 & \cellcolor{gray!20}22.500 & \cellcolor{gray!20}21.458 & \cellcolor{gray!20}6.855 & \cellcolor{gray!20}6.558 & \cellcolor{gray!20}113.423 & \cellcolor{gray!20}3.164 & \cellcolor{gray!20}2.922 & \cellcolor{gray!20}1.175 & \cellcolor{gray!20}1.163 & \cellcolor{gray!20}0.156 & \cellcolor{gray!20}0.151 & \cellcolor{gray!20}17.500 & \cellcolor{gray!20}16.259 \\

\multirow{2}{*}{News}
 & MGT & 4.423 & 4.312 & 0.872 & 2.321 & 12.624 & 66.000 & 43.518 & 63.647 & 3.435 & 439.956 & 61.577 & 39.069 & 15.454 & 11.048 & 1.786 & 1.618 & 1445.419 & 1004.841 \\
 &  HGT & \cellcolor{gray!20}13.030 & \cellcolor{gray!20}12.666 & \cellcolor{gray!20}3.172 & \cellcolor{gray!20}4.442 & \cellcolor{gray!20}30.974 & \cellcolor{gray!20}16.742 & \cellcolor{gray!20}16.331 & \cellcolor{gray!20}3.799 & \cellcolor{gray!20}5.157 & \cellcolor{gray!20}40.022 & \cellcolor{gray!20}3.712 & \cellcolor{gray!20}3.433 & \cellcolor{gray!20}1.297 & \cellcolor{gray!20}1.271 & \cellcolor{gray!20}0.254 & \cellcolor{gray!20}0.240 & \cellcolor{gray!20}29.733 & \cellcolor{gray!20}27.145 \\

\multirow{2}{*}{Poetry}
 & MGT & 7.306 & 6.775 & 3.087 & 1.969 & 41.181 & 131.272 & 116.825 & 128.017 & 2.694 & 1360.927 & 123.966 & 109.923 & 15.779 & 14.234 & 1.922 & 2.077 & 1477.864 & 1323.449 \\
 &  HGT & \cellcolor{gray!20}29.926 & \cellcolor{gray!20}26.064 & \cellcolor{gray!20}17.929 & \cellcolor{gray!20}1.263 & \cellcolor{gray!20}168.228 & \cellcolor{gray!20}86.721 & \cellcolor{gray!20}29.511 & \cellcolor{gray!20}219.813 & \cellcolor{gray!20}1.258 & \cellcolor{gray!20}2336.836 & \cellcolor{gray!20}56.795 & \cellcolor{gray!20}2.256 & \cellcolor{gray!20}3.040 & \cellcolor{gray!20}1.092 & \cellcolor{gray!20}0.323 & \cellcolor{gray!20}0.088 & \cellcolor{gray!20}203.953 & \cellcolor{gray!20}9.153 \\

\multirow{2}{*}{Recipes}
 & MGT & 2.353 & 2.272 & 0.510 & 1.513 & 10.274 & 11.255 & 10.289 & 5.772 & 2.308 & 64.101 & 8.902 & 7.975 & 4.719 & 4.329 & 1.403 & 1.427 & 371.932 & 332.902 \\
 &  HGT & \cellcolor{gray!20}7.292 & \cellcolor{gray!20}6.410 & \cellcolor{gray!20}3.446 & \cellcolor{gray!20}2.592 & \cellcolor{gray!20}38.450 & \cellcolor{gray!20}10.476 & \cellcolor{gray!20}8.723 & \cellcolor{gray!20}18.815 & \cellcolor{gray!20}3.681 & \cellcolor{gray!20}534.893 & \cellcolor{gray!20}3.183 & \cellcolor{gray!20}2.145 & \cellcolor{gray!20}1.455 & \cellcolor{gray!20}1.341 & \cellcolor{gray!20}0.309 & \cellcolor{gray!20}0.294 & \cellcolor{gray!20}45.546 & \cellcolor{gray!20}34.148 \\

\multirow{2}{*}{Reddit}
 & MGT & 5.919 & 5.167 & 5.425 & 1.537 & 162.881 & 14.161 & 7.275 & 67.858 & 1.787 & 1869.670 & 8.242 & 1.730 & 1.960 & 1.299 & 0.386 & 0.258 & 96.025 & 29.931 \\
 &  HGT & \cellcolor{gray!20}22.248 & \cellcolor{gray!20}19.196 & \cellcolor{gray!20}12.252 & \cellcolor{gray!20}1.067 & \cellcolor{gray!20}111.575 & \cellcolor{gray!20}28.612 & \cellcolor{gray!20}20.982 & \cellcolor{gray!20}44.944 & \cellcolor{gray!20}1.110 & \cellcolor{gray!20}590.347 & \cellcolor{gray!20}6.364 & \cellcolor{gray!20}1.340 & \cellcolor{gray!20}1.319 & \cellcolor{gray!20}1.073 & \cellcolor{gray!20}0.120 & \cellcolor{gray!20}0.070 & \cellcolor{gray!20}31.867 & \cellcolor{gray!20}7.270 \\

\multirow{2}{*}{Reviews}
 & MGT & 5.209 & 4.953 & 1.279 & 2.777 & 12.763 & 8.167 & 6.440 & 14.191 & 3.374 & 330.080 & 2.958 & 1.265 & 1.521 & 1.238 & 0.293 & 0.212 & 52.123 & 23.841 \\
 &  HGT & \cellcolor{gray!20}20.445 & \cellcolor{gray!20}17.844 & \cellcolor{gray!20}9.455 & \cellcolor{gray!20}6.658 & \cellcolor{gray!20}107.043 & \cellcolor{gray!20}39.051 & \cellcolor{gray!20}20.745 & \cellcolor{gray!20}196.345 & \cellcolor{gray!20}8.699 & \cellcolor{gray!20}4068.708 & \cellcolor{gray!20}18.606 & \cellcolor{gray!20}2.451 & \cellcolor{gray!20}1.750 & \cellcolor{gray!20}1.137 & \cellcolor{gray!20}0.175 & \cellcolor{gray!20}0.129 & \cellcolor{gray!20}75.012 & \cellcolor{gray!20}13.717 \\

\multirow{2}{*}{Wiki}
 & MGT & 3.814 & 3.600 & 1.952 & 1.704 & 50.395 & 7.619 & 5.483 & 13.342 & 2.174 & 369.460 & 3.805 & 1.646 & 2.045 & 1.438 & 0.498 & 0.357 & 104.545 & 43.780 \\
 &  HGT & \cellcolor{gray!20}11.641 & \cellcolor{gray!20}10.900 & \cellcolor{gray!20}4.590 & \cellcolor{gray!20}2.252 & \cellcolor{gray!20}41.626 & \cellcolor{gray!20}14.651 & \cellcolor{gray!20}13.829 & \cellcolor{gray!20}5.785 & \cellcolor{gray!20}3.431 & \cellcolor{gray!20}83.355 & \cellcolor{gray!20}3.010 & \cellcolor{gray!20}2.457 & \cellcolor{gray!20}1.305 & \cellcolor{gray!20}1.230 & \cellcolor{gray!20}0.235 & \cellcolor{gray!20}0.207 & \cellcolor{gray!20}30.543 & \cellcolor{gray!20}22.951 \\
 \midrule
\multirow{2}{*}{\textit{average}}
 & MGT & 
 4.906 &	4.595 &	2.015 &	2.050 &	41.963 &	31.805 &	25.375	 & 39.783 &	2.689 &	647.644 &	26.899 &	20.585 &	5.581 &	4.526 &	0.874 &	0.811 &	458.077 &	352.600\\
 &  HGT & \cellcolor{gray!20}16.956	 & \cellcolor{gray!20}15.286	 & \cellcolor{gray!20}7.682	 & \cellcolor{gray!20}3.289	 & \cellcolor{gray!20}80.847	 & \cellcolor{gray!20}29.283	 & \cellcolor{gray!20}18.296	 & \cellcolor{gray!20}62.723	 & \cellcolor{gray!20}4.048	 & \cellcolor{gray!20}977.289	 & \cellcolor{gray!20}12.327	 & \cellcolor{gray!20}2.549	 & \cellcolor{gray!20}1.585	 & \cellcolor{gray!20}1.201	 & \cellcolor{gray!20}0.231	 & \cellcolor{gray!20}0.180	 & \cellcolor{gray!20}58.464	 & \cellcolor{gray!20}20.090\\

\bottomrule
\end{tabular}
}
\caption{Main statistics of \mgt and \hgt perplexities and perplexity-features in RAID data, for each domain. \mgt statistics refer to the average over all generators.
}   
\label{tab-app:text_stats}
\end{sidewaystable*}

\clearpage
 \begin{sidewaystable*}[t!]
\centering
\scalebox{0.8}{
\setlength{\tabcolsep}{3pt}
\renewcommand{\arraystretch}{1.15}
\begin{tabular}{l|l|ccccc|ccccc|cc|cc|cc|cc} 
\toprule
Domain & Type & Mean & Median &	Stdev &Min &Max  &	Mean& Median & Stdev 	 & Min & Max & Mean & Median	& Mean & Median 	& Mean 	& Median  	& Mean& Median \\
& & 
$ppl$ & $ppl$ &	$ppl$ & $ppl$ &	$ppl$ & $ppl_{shuf}$ & $ppl_{shuf}$ &	$ppl_{shuf}$ & $ppl_{shuf}$ &	$ppl_{shuf}$ & diff & diff & ratio  & ratio & logratio & logratio & change & change \\
\midrule

\multirow{2}{*}{Abstracts}	 & 	MGT	 & 	8.479	 & 	7.686	 & 	3.808	 & 	2.736	 & 	60.599	 & 	13.198	 & 	10.702	 & 	22.941	 & 	3.475	 & 	662.957	 & 	4.719	 & 	2.536	 & 	1.616	 & 	1.327	 & 	0.364	 & 	0.282	 & 	61.559	 & 	32.728	 \\ 
	 & 	HGT	 & \cellcolor{gray!20}26.345	 & \cellcolor{gray!20}24.814	 & \cellcolor{gray!20}9.238	 & \cellcolor{gray!20}4.612	 & \cellcolor{gray!20}75.626	 & \cellcolor{gray!20}32.753	 & \cellcolor{gray!20}31.251	 & \cellcolor{gray!20}10.950	 & \cellcolor{gray!20}4.751	 & \cellcolor{gray!20}87.688	 & \cellcolor{gray!20}6.408	 & \cellcolor{gray!20}5.557	 & \cellcolor{gray!20}1.257	 & \cellcolor{gray!20}1.224	 & \cellcolor{gray!20}0.219	 & \cellcolor{gray!20}0.202	 & \cellcolor{gray!20}25.701	 & \cellcolor{gray!20}22.401	 \\ 
\multirow{2}{*}{Books}	 & 	MGT	 & 	6.387	 & 	6.084	 & 	1.689	 & 	2.896	 & 	22.905	 & 	10.804	 & 	8.760	 & 	16.631	 & 	3.855	 & 	485.647	 & 	4.417	 & 	2.437	 & 	1.784	 & 	1.394	 & 	0.420	 & 	0.329	 & 	78.414	 & 	39.379	 \\ 
	 & 	HGT	 & \cellcolor{gray!20}23.545	 & \cellcolor{gray!20}22.112	 & \cellcolor{gray!20}7.993	 & \cellcolor{gray!20}6.423	 & \cellcolor{gray!20}133.197	 & \cellcolor{gray!20}27.511	 & \cellcolor{gray!20}26.035	 & \cellcolor{gray!20}8.865	 & \cellcolor{gray!20}8.854	 & \cellcolor{gray!20}146.791	 & \cellcolor{gray!20}3.966	 & \cellcolor{gray!20}3.549	 & \cellcolor{gray!20}1.182	 & \cellcolor{gray!20}1.161	 & \cellcolor{gray!20}0.159	 & \cellcolor{gray!20}0.149	 & \cellcolor{gray!20}18.222	 & \cellcolor{gray!20}16.059	 \\ 
\multirow{2}{*}{News}	 & 	MGT	 & 	5.066	 & 	4.873	 & 	1.267	 & 	2.413	 & 	20.369	 & 	88.458	 & 	57.741	 & 	87.200	 & 	3.782	 & 	653.288	 & 	83.392	 & 	52.646	 & 	19.298	 & 	13.690	 & 	1.919	 & 	1.750	 & 	1829.770	 & 	1269.010	 \\ 
	 & 	HGT	 & \cellcolor{gray!20}15.288	 & \cellcolor{gray!20}14.682	 & \cellcolor{gray!20}3.973	 & \cellcolor{gray!20}2.378	 & \cellcolor{gray!20}46.755	 & \cellcolor{gray!20}19.340	 & \cellcolor{gray!20}18.693	 & \cellcolor{gray!20}4.612	 & \cellcolor{gray!20}6.159	 & \cellcolor{gray!20}52.038	 & \cellcolor{gray!20}4.053	 & \cellcolor{gray!20}3.671	 & \cellcolor{gray!20}1.284	 & \cellcolor{gray!20}1.250	 & \cellcolor{gray!20}0.239	 & \cellcolor{gray!20}0.223	 & \cellcolor{gray!20}28.352	 & \cellcolor{gray!20}25.026	 \\ 
\multirow{2}{*}{Poetry}	 & 	MGT	 & 	15.427	 & 	13.617	 & 	10.653	 & 	2.594	 & 	196.132	 & 	164.876	 & 	147.213	 & 	151.566	 & 	3.061	 & 	1470.259	 & 	149.449	 & 	132.616	 & 	10.636	 & 	9.157	 & 	1.733	 & 	1.849	 & 	963.582	 & 	815.650	 \\ 
	 & 	HGT	 & \cellcolor{gray!20}40.102	 & \cellcolor{gray!20}35.147	 & \cellcolor{gray!20}22.961	 & \cellcolor{gray!20}1.164	 & \cellcolor{gray!20}224.335	 & \cellcolor{gray!20}107.678	 & \cellcolor{gray!20}39.224	 & \cellcolor{gray!20}262.881	 & \cellcolor{gray!20}1.335	 & \cellcolor{gray!20}2830.231	 & \cellcolor{gray!20}67.575	 & \cellcolor{gray!20}2.043	 & \cellcolor{gray!20}2.963	 & \cellcolor{gray!20}1.062	 & \cellcolor{gray!20}0.293	 & \cellcolor{gray!20}0.060	 & \cellcolor{gray!20}196.255	 & \cellcolor{gray!20}6.188	 \\ 
\multirow{2}{*}{Recipes}	 & 	MGT	 & 	3.212	 & 	2.964	 & 	1.718	 & 	1.708	 & 	33.943	 & 	14.559	 & 	13.141	 & 	7.773	 & 	2.772	 & 	89.993	 & 	11.347	 & 	10.094	 & 	4.634	 & 	4.265	 & 	1.405	 & 	1.429	 & 	363.374	 & 	326.502	 \\ 
	 & 	HGT	 & \cellcolor{gray!20}9.435	 & \cellcolor{gray!20}8.444	 & \cellcolor{gray!20}4.290	 & \cellcolor{gray!20}2.625	 & \cellcolor{gray!20}46.113	 & \cellcolor{gray!20}12.820	 & \cellcolor{gray!20}10.680	 & \cellcolor{gray!20}22.865	 & \cellcolor{gray!20}4.112	 & \cellcolor{gray!20}630.171	 & \cellcolor{gray!20}3.385	 & \cellcolor{gray!20}2.034	 & \cellcolor{gray!20}1.368	 & \cellcolor{gray!20}1.251	 & \cellcolor{gray!20}0.244	 & \cellcolor{gray!20}0.224	 & \cellcolor{gray!20}36.759	 & \cellcolor{gray!20}25.109	 \\ 
\multirow{2}{*}{Reddit}	 & 	MGT	 & 	7.225	 & 	6.083	 & 	7.543	 & 	1.881	 & 	184.676	 & 	18.867	 & 	8.891	 & 	108.309	 & 	2.625	 & 	2908.624	 & 	11.642	 & 	2.340	 & 	2.622	 & 	1.342	 & 	0.429	 & 	0.289	 & 	162.228	 & 	34.218	 \\ 
	 & 	HGT	 & \cellcolor{gray!20}28.399	 & \cellcolor{gray!20}24.665	 & \cellcolor{gray!20}15.366	 & \cellcolor{gray!20}1.068	 & \cellcolor{gray!20}189.922	 & \cellcolor{gray!20}35.973	 & \cellcolor{gray!20}26.369	 & \cellcolor{gray!20}55.827	 & \cellcolor{gray!20}1.078	 & \cellcolor{gray!20}733.526	 & \cellcolor{gray!20}7.574	 & \cellcolor{gray!20}1.444	 & \cellcolor{gray!20}1.310	 & \cellcolor{gray!20}1.061	 & \cellcolor{gray!20}0.105	 & \cellcolor{gray!20}0.059	 & \cellcolor{gray!20}31.045	 & \cellcolor{gray!20}6.100	 \\ 
\multirow{2}{*}{Reviews}	 & 	MGT	 & 	5.991	 & 	5.693	 & 	1.541	 & 	3.041	 & 	14.859	 & 	10.018	 & 	7.665	 & 	20.138	 & 	3.777	 & 	470.756	 & 	4.027	 & 	1.732	 & 	1.625	 & 	1.279	 & 	0.331	 & 	0.244	 & 	62.543	 & 	27.931	 \\ 
	 & 	HGT	 & \cellcolor{gray!20}23.562	 & \cellcolor{gray!20}20.606	 & \cellcolor{gray!20}11.351	 & \cellcolor{gray!20}7.505	 & \cellcolor{gray!20}120.985	 & \cellcolor{gray!20}45.022	 & \cellcolor{gray!20}23.592	 & \cellcolor{gray!20}224.510	 & \cellcolor{gray!20}7.091	 & \cellcolor{gray!20}4664.978	 & \cellcolor{gray!20}21.460	 & \cellcolor{gray!20}2.552	 & \cellcolor{gray!20}1.846	 & \cellcolor{gray!20}1.125	 & \cellcolor{gray!20}0.169	 & \cellcolor{gray!20}0.118	 & \cellcolor{gray!20}84.551	 & \cellcolor{gray!20}12.481	 \\ 
\multirow{2}{*}{Wiki}	 & 	MGT	 & 	5.004	 & 	4.614	 & 	2.195	 & 	2.013	 & 	45.196	 & 	10.638	 & 	7.463	 & 	16.656	 & 	2.613	 & 	416.953	 & 	5.634	 & 	2.544	 & 	2.191	 & 	1.522	 & 	0.550	 & 	0.414	 & 	119.117	 & 	52.215	 \\ 
	 & 	HGT	 & \cellcolor{gray!20}18.080	 & \cellcolor{gray!20}16.708	 & \cellcolor{gray!20}7.138	 & \cellcolor{gray!20}4.180	 & \cellcolor{gray!20}65.997	 & \cellcolor{gray!20}21.738	 & \cellcolor{gray!20}20.368	 & \cellcolor{gray!20}8.466	 & \cellcolor{gray!20}5.180	 & \cellcolor{gray!20}124.496	 & \cellcolor{gray!20}3.658	 & \cellcolor{gray!20}2.849	 & \cellcolor{gray!20}1.242	 & \cellcolor{gray!20}1.172	 & \cellcolor{gray!20}0.190	 & \cellcolor{gray!20}0.159	 & \cellcolor{gray!20}24.180	 & \cellcolor{gray!20}17.200	 \\ 

\midrule
\multirow{2}{*}{\textit{average}}	 & 	MGT	 & 	7.099	 & 	6.452	 & 	3.802	 & 	2.410	 & 	72.335	 & 	41.427	 & 	32.697	 & 	53.902	 & 	3.245	 & 	894.810	 & 	34.328	 & 	25.868	 & 	5.551	 & 	4.247	 & 	0.894	 & 	0.823	 & 	455.073	 & 	324.704	 \\ 
	 & 	HGT	 & 	\cellcolor{gray!20}23.094	 & 	\cellcolor{gray!20}20.897	 & \cellcolor{gray!20}10.289	 & \cellcolor{gray!20}3.744	 & \cellcolor{gray!20}112.866	 & \cellcolor{gray!20}37.854	 & \cellcolor{gray!20}24.527	 & \cellcolor{gray!20}74.872	 & \cellcolor{gray!20}4.820	 & \cellcolor{gray!20}1158.740	 & \cellcolor{gray!20}14.760	 & \cellcolor{gray!20}2.962	 & \cellcolor{gray!20}1.556	 & \cellcolor{gray!20}1.163	 & \cellcolor{gray!20}0.202	 & \cellcolor{gray!20}0.149	 & \cellcolor{gray!20}55.633	 & \cellcolor{gray!20}16.321	 \\

\bottomrule
\end{tabular}
}
\caption{Main statistics of \mgt and \hgt perplexities and perplexity-features in RAID data, for each domain, using GPT-2 as perplexity model. \mgt statistics refer to the average over all generators.}
\label{tab-app:text_stats_gpt2ppl}
\end{sidewaystable*}

%% file: appendix_material/feature_sig.tex
\begin{table*}[t]
\centering
\scalebox{0.85}{
\setlength{\tabcolsep}{6pt}
\renewcommand{\arraystretch}{1.15}
\begin{tabular}{l|l|rrrr|rrrr} 
\toprule
Domain & Feature & \multicolumn{4}{c}{Welch} & \multicolumn{4}{c}{Mann-Whitney} \\
 &  & effect & CI low & CI high & p-value & effect & CI low & CI high & p-value \\
\midrule 
\multirow{5}{*}{Abstracts} & 
diff & -1.296 & -2.233 & -0.360 & 6.71E-03 & 0.654 & 0.626 & 0.681 & 3.59E-248\\
& sum & -15.227 & -16.292 & -14.161 & 2.49E-150 & 0.927 & 0.913 & 0.939 & 0.00E+00\\
& ratio & 0.110 & 0.022 & 0.198 & 1.41E-02 & 0.007 & -0.032 & 0.046 & \cellcolor{gray!20}7.05E-01\\
& logratio & 0.013 & -0.001 & 0.026 & 6.40E-02 & 0.007 & -0.030 & 0.045 & \cellcolor{gray!20}7.05E-01\\
& change & 0.110 & 0.022 & 0.198 & 1.41E-02 & 0.007 & -0.031 & 0.045 & \cellcolor{gray!20}7.05E-01\\
  \midrule
\multirow{5}{*}{Books} & diff & 4.207 & 2.829 & 5.585 & 2.59E-09 & -0.057 & -0.096 & -0.018 & 3.01E-03\\
& sum & -21.769 & -23.286 & -20.252 & 6.70E-151 & 0.897 & 0.880 & 0.913 & 0.00E+00\\
& ratio & 1.012 & 0.833 & 1.190 & 7.21E-28 & -0.717 & -0.742 & -0.691 & 3.79E-300\\
& logratio & 0.397 & 0.373 & 0.421 & 4.22E-184 & -0.717 & -0.743 & -0.690 & 3.79E-300\\
& change & 1.012 & 0.833 & 1.190 & 7.21E-28 & -0.717 & -0.742 & -0.690 & 3.79E-300\\
  \midrule
\multirow{5}{*}{News} & diff & 184.313 & 178.657 & 189.969 & 0.00E+00 & -0.982 & -0.989 & -0.975 & 0.00E+00\\
& sum & 167.667 & 161.960 & 173.374 & 0.00E+00 & -0.941 & -0.955 & -0.925 & 0.00E+00\\
& ratio & 39.300 & 38.262 & 40.338 & 0.00E+00 & -0.995 & -0.998 & -0.991 & 0.00E+00\\
& logratio & 3.260 & 3.227 & 3.293 & 0.00E+00 & -0.995 & -0.998 & -0.991 & 0.00E+00\\
& change & 39.300 & 38.262 & 40.338 & 0.00E+00 & -0.995 & -0.998 & -0.991 & 0.00E+00\\
  \midrule
\multirow{5}{*}{Poetry} & diff & 85.560 & 70.825 & 100.294 & 1.62E-29 & -0.788 & -0.814 & -0.762 & 0.00E+00\\
& sum & 38.446 & 23.251 & 53.642 & 7.35E-07 & -0.256 & -0.293 & -0.218 & 1.22E-39\\
& ratio & 19.918 & 18.794 & 21.043 & 1.52E-214 & -0.879 & -0.897 & -0.860 & 0.00E+00\\
& logratio & 2.381 & 2.321 & 2.442 & 0.00E+00 & -0.879 & -0.896 & -0.861 & 0.00E+00\\
& change & 19.918 & 18.794 & 21.043 & 1.52E-214 & -0.879 & -0.897 & -0.860 & 0.00E+00\\
  \midrule
\multirow{5}{*}{Recipes} & diff & 4.915 & 4.128 & 5.702 & 1.49E-33 & -0.776 & -0.799 & -0.752 & 0.00E+00\\
& sum & -4.920 & -5.789 & -4.051 & 6.20E-28 & 0.455 & 0.421 & 0.488 & 1.43E-121\\
& ratio & 2.847 & 2.700 & 2.994 & 7.46E-260 & -0.965 & -0.973 & -0.956 & 0.00E+00\\
& logratio & 1.012 & 0.987 & 1.036 & 0.00E+00 & -0.965 & -0.974 & -0.956 & 0.00E+00\\
& change & 2.847 & 2.700 & 2.994 & 7.46E-260 & -0.965 & -0.974 & -0.956 & 0.00E+00\\
  \midrule
\multirow{5}{*}{Reddit} & diff & 24.896 & 10.822 & 38.969 & 5.33E-04 & -0.326 & -0.360 & -0.294 & 1.12E-63\\
& sum & -3.545 & -18.495 & 11.406 & \cellcolor{gray!20}6.42E-01 & 0.744 & 0.717 & 0.770 & 0.00E+00\\
& ratio & 2.146 & 1.608 & 2.684 & 8.39E-15 & -0.707 & -0.733 & -0.681 & 3.07E-292\\
& logratio & 0.492 & 0.453 & 0.532 & 2.17E-119 & -0.707 & -0.734 & -0.681 & 3.07E-292\\
& change & 2.146 & 1.608 & 2.684 & 8.39E-15 & -0.707 & -0.733 & -0.681 & 3.07E-292\\
  \midrule
\multirow{5}{*}{Reviews} &diff & -21.663 & -39.635 & -3.691 & 1.82E-02 & 0.045 & -0.008 & 0.100 & \cellcolor{gray!20}9.01E-02\\
& sum & -51.622 & -69.793 & -33.451 & 3.20E-08 & 0.899 & 0.875 & 0.922 & 1.31E-250\\
& ratio & 0.248 & -0.427 & 0.923 & \cellcolor{gray!20}4.72E-01 & -0.608 & -0.647 & -0.568 & 9.79E-116\\
& logratio & 0.310 & 0.262 & 0.359 & 1.70E-34 & -0.608 & -0.647 & -0.568 & 9.79E-116\\
& change & 0.248 & -0.427 & 0.923 & \cellcolor{gray!20}4.72E-01 & -0.608 & -0.649 & -0.568 & 9.79E-116\\
  \midrule
\multirow{5}{*}{Wiki} & diff & 7.683 & 5.815 & 9.552 & 1.33E-15 & -0.182 & -0.221 & -0.144 & 4.53E-21\\
& sum & -7.672 & -9.612 & -5.733 & 1.37E-14 & 0.673 & 0.645 & 0.702 & 6.25E-265\\
& ratio & 2.341 & 1.948 & 2.734 & 2.16E-30 & -0.687 & -0.711 & -0.660 & 1.95E-275\\
& logratio & 0.600 & 0.565 & 0.635 & 5.85E-198 & -0.687 & -0.712 & -0.661 & 1.95E-275\\
& change & 2.341 & 1.948 & 2.734 & 2.16E-30 & -0.687 & -0.711 & -0.661 & 1.95E-275\\
\bottomrule
\end{tabular}
}
\caption{Statistical significance   of perplexity-features in RAID data: Effect sizes, confidence intervals, and p-values  for the Welch’s t-tests and the Mann-Whitney U tests. Grey cells  denote p-values $\geq 0.05$ (i.e., the null hypothesis cannot be rejected).} 
\label{tab-app:feat_sig}
\end{table*}

\begin{table*}[t]
\centering
\scalebox{0.85}{
\setlength{\tabcolsep}{6pt}
\renewcommand{\arraystretch}{1.15}
\begin{tabular}{l|l|rrrr|rrrr} 
\toprule
Domain & Feature & \multicolumn{4}{c}{Welch} & \multicolumn{4}{c}{Mann-Whitney} \\
 &  & effect & CI low & CI high & p-value & effect & CI low & CI high & p-value \\
\midrule \multirow{5}{*}{Abstracts}	 &  	diff	 &  	-1.512	 &  	-2.679	 &  	-0.344	 &  	1.12E-02	 &  	0.451	 &  	0.417	 &  	0.485	 &  	3.21E-119	  \\
  	 &  	sum	 &  	-34.170	 &  	-35.861	 &  	-32.480	 &  	1.11E-276	 &  	0.907	 &  	0.892	 &  	0.922	 &  	0.00E+00	  \\
  	 &  	ratio	 &  	0.180	 &  	0.139	 &  	0.221	 &  	1.58E-17	 &  	-0.401	 &  	-0.436	 &  	-0.366	 &  	9.60E-95	  \\
   	 &  	logratio	 &  	0.104	 &  	0.092	 &  	0.115	 &  	5.71E-67	 &  	-0.401	 &  	-0.435	 &  	-0.367	 &  	9.60E-95	  \\
   	 &  	change	 &  	0.180	 &  	0.139	 &  	0.221	 &  	1.58E-17	 &  	-0.401	 &  	-0.435	 &  	-0.367	 &  	9.60E-95	  \\
\midrule \multirow{5}{*}{Books}	 &  	diff	 &  	6.555	 &  	4.692	 &  	8.419	 &  	7.24E-12	 &  	-0.214	 &  	-0.252	 &  	-0.177	 &  	1.67E-28	  \\
   	 &  	sum	 &  	-25.021	 &  	-27.064	 &  	-22.978	 &  	2.96E-114	 &  	0.863	 &  	0.843	 &  	0.881	 &  	0.00E+00	  \\
   	 &  	ratio	 &  	1.175	 &  	0.977	 &  	1.373	 &  	2.89E-30	 &  	-0.824	 &  	-0.843	 &  	-0.804	 &  	0.00E+00	  \\
   	 &  	logratio	 &  	0.465	 &  	0.440	 &  	0.490	 &  	8.96E-230	 &  	-0.824	 &  	-0.844	 &  	-0.804	 &  	0.00E+00	  \\
   	 &  	change	 &  	1.175	 &  	0.977	 &  	1.373	 &  	2.89E-30	 &  	-0.824	 &  	-0.843	 &  	-0.805	 &  	0.00E+00	  \\
\midrule \multirow{5}{*}{News}	 &  	diff	 &  	248.104	 &  	240.264	 &  	255.945	 &  	0.00E+00	 &  	-0.986	 &  	-0.992	 &  	-0.979	 &  	0.00E+00	  \\
   	 &  	sum	 &  	228.088	 &  	220.192	 &  	235.985	 &  	0.00E+00	 &  	-0.946	 &  	-0.960	 &  	-0.932	 &  	0.00E+00	  \\
   	 &  	ratio	 &  	47.821	 &  	46.488	 &  	49.155	 &  	0.00E+00	 &  	-0.995	 &  	-0.998	 &  	-0.992	 &  	0.00E+00	  \\
   	 &  	logratio	 &  	3.448	 &  	3.413	 &  	3.483	 &  	0.00E+00	 &  	-0.995	 &  	-0.998	 &  	-0.992	 &  	0.00E+00	  \\
   	 &  	change	 &  	47.821	 &  	46.488	 &  	49.155	 &  	0.00E+00	 &  	-0.995	 &  	-0.998	 &  	-0.992	 &  	0.00E+00	  \\
\midrule \multirow{5}{*}{Poetry}	 &  	diff	 &  	104.325	 &  	87.065	 &  	121.584	 &  	8.44E-32	 &  	-0.778	 &  	-0.805	 &  	-0.751	 &  	0.00E+00	  \\
   	 &  	sum	 &  	48.515	 &  	30.372	 &  	66.658	 &  	1.67E-07	 &  	-0.230	 &  	-0.268	 &  	-0.192	 &  	1.92E-32	  \\
   	 &  	ratio	 &  	12.767	 &  	12.071	 &  	13.462	 &  	1.07E-240	 &  	-0.828	 &  	-0.851	 &  	-0.805	 &  	0.00E+00	  \\
   	 &  	logratio	 &  	2.183	 &  	2.128	 &  	2.238	 &  	0.00E+00	 &  	-0.828	 &  	-0.850	 &  	-0.804	 &  	0.00E+00	  \\
   	 &  	change	 &  	12.767	 &  	12.071	 &  	13.462	 &  	1.07E-240	 &  	-0.828	 &  	-0.851	 &  	-0.804	 &  	0.00E+00	  \\
\midrule \multirow{5}{*}{Recipes}	 &  	diff	 &  	7.226	 &  	6.212	 &  	8.240	 &  	9.54E-43	 &  	-0.845	 &  	-0.866	 &  	-0.825	 &  	0.00E+00	  \\
   	 &  	sum	 &  	-5.273	 &  	-6.397	 &  	-4.149	 &  	7.38E-20	 &  	0.392	 &  	0.357	 &  	0.426	 &  	1.06E-90	  \\
   	 &  	ratio	 &  	2.886	 &  	2.752	 &  	3.019	 &  	0.00E-02	 &  	-0.958	 &  	-0.969	 &  	-0.948	 &  	0.00E+00	  \\
   	 &  	logratio	 &  	1.094	 &  	1.070	 &  	1.117	 &  	0.00E+00	 &  	-0.958	 &  	-0.968	 &  	-0.947	 &  	0.00E+00	  \\
   	 &  	change	 &  	2.886	 &  	2.752	 &  	3.019	 &  	0.00E-02	 &  	-0.958	 &  	-0.968	 &  	-0.947	 &  	0.00E+00	  \\
\midrule \multirow{5}{*}{Reddit}	 &  	diff	 &  	29.345	 &  	15.058	 &  	43.632	 &  	5.84E-05	 &  	-0.517	 &  	-0.548	 &  	-0.488	 &  	4.66E-157	  \\
   	 &  	sum	 &  	-7.763	 &  	-24.028	 &  	8.502	 &  	\cellcolor{gray!20}3.49E-01	 &  	0.752	 &  	0.725	 &  	0.778	 &  	0.00E+00	  \\
   	 &  	ratio	 &  	2.416	 &  	1.909	 &  	2.923	 &  	2.48E-20	 &  	-0.877	 &  	-0.894	 &  	-0.859	 &  	0.00E+00	  \\
   	 &  	logratio	 &  	0.588	 &  	0.548	 &  	0.627	 &  	6.14E-161	 &  	-0.877	 &  	-0.894	 &  	-0.859	 &  	0.00E+00	  \\
   	 &  	change	 &  	2.416	 &  	1.909	 &  	2.923	 &  	2.48E-20	 &  	-0.877	 &  	-0.895	 &  	-0.860	 &  	0.00E+00	  \\
\midrule \multirow{5}{*}{Reviews}	 &  	diff	 &  	-22.476	 &  	-42.116	 &  	-2.836	 &  	2.49E-02	 &  	-0.120	 &  	-0.172	 &  	-0.066	 &  	6.70E-06	  \\
   	 &  	sum	 &  	-56.704	 &  	-76.511	 &  	-36.896	 &  	2.51E-08	 &  	0.875	 &  	0.848	 &  	0.900	 &  	2.25E-237	  \\
   	 &  	ratio	 &  	0.384	 &  	-0.384	 &  	1.152	 &  	\cellcolor{gray!20}3.27E-01	 &  	-0.736	 &  	-0.767	 &  	-0.704	 &  	1.89E-168	  \\
   	 &  	logratio	 &  	0.370	 &  	0.320	 &  	0.421	 &  	1.76E-44	 &  	-0.736	 &  	-0.768	 &  	-0.702	 &  	1.89E-168	  \\
   	 &  	change	 &  	0.384	 &  	-0.384	 &  	1.152	 &  	\cellcolor{gray!20}3.27E-01	 &  	-0.736	 &  	-0.770	 &  	-0.703	 &  	1.89E-168	  \\
\midrule \multirow{5}{*}{Wiki}	 &  	diff	 &  	12.661	 &  	9.844	 &  	15.479	 &  	2.80E-18	 &  	-0.368	 &  	-0.403	 &  	-0.332	 &  	2.37E-80	  \\
   	 &  	sum	 &  	-11.973	 &  	-14.917	 &  	-9.029	 &  	2.52E-15	 &  	0.676	 &  	0.648	 &  	0.704	 &  	4.37E-267	  \\
   	 &  	ratio	 &  	2.423	 &  	2.080	 &  	2.766	 &  	1.58E-41	 &  	-0.834	 &  	-0.850	 &  	-0.815	 &  	0.00E+00	  \\
   	 &  	logratio	 &  	0.687	 &  	0.653	 &  	0.721	 &  	3.81E-253	 &  	-0.834	 &  	-0.851	 &  	-0.816	 &  	0.00E+00	  \\
   	 &  	change	 &  	2.423	 &  	2.080	 &  	2.766	 &  	1.58E-41	 &  	-0.834	 &  	-0.851	 &  	-0.816	 &  	0.00E+00	  \\
\bottomrule
\end{tabular}
}
\caption{Statistical significance   of perplexity-features in RAID data, for each domain, using GPT-2 as perplexity model: Effect sizes, confidence intervals, and p-values  for the Welch’s t-tests and the Mann-Whitney U tests. Grey cells denote p-values $\geq 0.05$ (i.e., the null hypothesis cannot be rejected).}
\label{tab-app:feat_sig_gpt2ppl}
\end{table*}